\documentclass[utf8]{FrontiersinHarvard} 
\usepackage{url,hyperref,lineno,microtype,subcaption}
\usepackage[onehalfspacing]{setspace}

\usepackage{graphicx}%
\usepackage{multirow}%
\usepackage{amsmath,amssymb,amsfonts}%
\usepackage{amsthm}%
\usepackage{mathrsfs}%
\usepackage[title]{appendix}%
\usepackage{xcolor}%
\usepackage{textcomp}%
\usepackage{manyfoot}%
\usepackage{booktabs}%
\usepackage{algorithm}%
\usepackage{algorithmicx}%
\usepackage{algpseudocode}%
\usepackage{listings}%
\usepackage{tabularx}

\usepackage{verbatim} 

\usepackage{caption}
\usepackage{subcaption}

\usepackage{csquotes}

\usepackage{xspace}

\newtheorem{assumption}{Assumption}

\usepackage[normalem]{ulem}
\usepackage[commandnameprefix=always]{changes}

\definecolor{runtimecolor}{HTML}{8A3F80}
\definecolor{entitycolor}{HTML}{5B78A1}
\definecolor{relationcolor}{HTML}{FFF6E3}
\definecolor{attributecolor}{HTML}{82B366}

\newcommand\acronym{ROSA\xspace}

\newcommand\runtime[1]{{\textcolor{runtimecolor}{#1}}}
\newcommand\runtimefont{\runtime{purple font\xspace}\xspace}

\newcommand\entity[1]{\texttt{{\small{#1}}}}
\newcommand\instance[1]{\textit{{#1}}}
\newcommand\relationship[1]{\texttt{{\small{#1}}}}
\newcommand\attribute[1]{\texttt{{\small{#1}}}}

\usepackage{listings}
\definecolor{codegreen}{rgb}{0,0.6,0}
\definecolor{codegray}{rgb}{0.5,0.5,0.5}
\definecolor{codepurple}{rgb}{0.58,0,0.82}
\definecolor{backcolour}{rgb}{0.95,0.95,0.92}

\definecolor{runtime}{HTML}{914800}
 
\lstdefinestyle{mystyle}{
    frame=single,
    commentstyle=\color{blue}\sl\sffamily,
    morecomment=[l]{\#},
    keywordstyle=\color{purple},
    numberstyle=\tiny\color{codegray},
    stringstyle=\color{olive},
    basicstyle=\sffamily\scriptsize,
    breakatwhitespace=false,         
    breaklines=true,                 
    captionpos=b,                    
    keepspaces=true,                 
    numbers=left,                    
    numbersep=5pt,                  
    showspaces=false,                
    showstringspaces=false,
    showtabs=false,                  
    tabsize=2,
    literate={\{}{{\color{purple}{\{}}}1 {\}}{{\color{purple}{\}}}}1
}
\lstdefinelanguage{PDDL}
{
  sensitive=false,    
  morecomment=[l]{;}, 
  alsoletter={:,-,?},   
  keywords=[1]{
    define,domain,problem,when,forall,exists,either,
    :domain,:requirements,:types,:objects,:constants,
    :predicates,:action,:parameters,:precondition,:effect,
    :fluents,:primary-effect,:side-effect,:init,:goal,
    :strips,:adl,:equality,:typing,:conditional-effects,
    :negative-preconditions,:disjunctive-preconditions,
    :existential-preconditions,:universal-preconditions,:quantified-preconditions,
    :functions,assign,increase,decrease,scale-up,scale-down,
    :metric,minimize,maximize,
    :durative-actions,:duration-inequalities,:continuous-effects,
    :durative-action,:duration,:condition,
  }
  keywordstyle=[1]\color{purple},
  keywords=[2]{and, or, over,all, at, end, not},
  keywordstyle=[2]\color{teal},
  keywordsprefix=?
}

\lstdefinelanguage{TypeQL}
{
  sensitive=yes,    
  morecomment=[l]{\#}, 
  alsoletter={:},   
  keywords=[1]{
    entity, attribute, relation, 
  },
  keywordstyle=[1]\color{teal},
  keywords=[2]{@key},
  keywordstyle=[2]\color{orange},
  keywords=[3]{
    sub, abstract, owns, value, relates, plays,
    rule, when, not, has, then, isa, \},\{,
    fetch, match, insert, define, :,
  },
  keywordstyle=[3]\color{purple},
  morestring=[b]{'},
  keywordsprefix=\$
}

\usepackage{cleveref}
\crefformat{footnote}{#2\footnotemark[#1]#3}

\def\keyFont{\fontsize{8}{11}\helveticabold }
\def\firstAuthorLast{Silva {et~al.}} 
\def\Authors{Gustavo Rezende Silva\,$^{1,*}$, Juliane P{\"a}{\ss}ler\,$^{2}$, S. Lizeth Tapia Tarifa\,$^{2}$, Einar Broch Johnsen\,$^{2}$,  and Carlos Hern{\'a}ndez Corbato\,$^{1}$}

\def\Address{$^{1}$Delft University of Technology, Delft, Netherlands \\
$^{2}$University of Oslo, Oslo, Norway  }

\def\corrAuthor{Gustavo Rezende Silva}

\def\corrEmail{g.rezendesilva@tudelft.nl}

\begin{document}
\onecolumn
\firstpage{1}

\title[\acronym: A Knowledge-based Solution for Robot Self-Adaptation]{\acronym: A Knowledge-based Solution for Robot Self-Adaptation} 

\author[\firstAuthorLast ]{\Authors} 
\address{} 
\correspondance{} 

\extraAuth{}

\maketitle

\begin{abstract}
Autonomous robots must operate in diverse environments and handle multiple tasks despite uncertainties. This creates challenges in designing software architectures and task decision-making algorithms, as different contexts may require distinct task logic and architectural configurations. To address this, robotic systems can be designed as self-adaptive systems capable of adapting their task execution and software architecture at runtime based on their context.
This paper introduces ROSA, a novel knowledge-based framework for RObot Self-Adaptation, which enables task-and-architecture co-adaptation (TACA) in robotic systems. ROSA achieves this by providing a knowledge model that captures all application-specific knowledge required for adaptation and by reasoning over this knowledge at runtime to determine when and how adaptation should occur.
In addition to a conceptual framework, this work provides an open-source ROS 2-based reference implementation of ROSA and evaluates its feasibility and performance in an underwater robotics application. Experimental results highlight ROSA’s advantages in reusability and development effort for designing self-adaptive robotic systems.

\tiny
 \keyFont{ \section{Keywords:} self-adaptation, knowledge representation, underwater vehicle, robotics}
\end{abstract}

\section{Introduction}

A current challenge in robotics is designing software architectures and task decision-making algorithms that enable robots to autonomously perform multiple tasks in diverse environments while handling internal and environmental uncertainties. This challenge arises because different contexts may demand distinct task logic and architectural configurations. At runtime, certain actions may become unfeasible, requiring the robot to adapt its task execution to ensure mission completion. For example, a robot navigating through an environment might run out of battery during its operation, requiring it to adapt its task execution to include a recharge action. Additionally, actions may require different architectural configurations depending on the context. For example, a navigation action that relies on vision-based localization cannot be executed in environments without lights but could potentially be executed with an alternative architectural configuration that employs a localization strategy based on lidar. This becomes even more challenging when both the robot’s task execution and its architectural configuration need to be adapted. For instance, when a robot runs out of battery while navigating, it must simultaneously adapt its architecture to a configuration that consumes less energy and its task execution to include a recharge action and to navigate along paths that are better suited to the new configuration. To address this challenge, robots can be designed as self-adaptive systems~(SASs) with the ability to perform \emph{task-and-architecture co-adaptation}~(TACA)~\citep{ task_co_adapt_camara_garlan}, i.e., simultaneously adapt their task execution and software architecture dependently during runtime. This work focuses on proposing a systematic solution for enabling TACA that can be reused with different robotic systems.

A common approach to enable self-adaptation in software systems is to design them as two-layered systems containing a managing and managed subsystem~\citep{weyns2020introduction}, where the managing subsystem monitors and reconfigures the managed subsystem, and the managed subsystem is responsible for the domain logic. This design facilitates the development and maintenance of the system by creating a clear separation between the adaptation and the domain logic. While several solutions have been proposed for solving either architectural~\citep{ALBERTS2025112258} or task adaptation in robotic systems~\citep{carreno2021situation, hamilton2022towards}, there are some works that partially address TACA~\mbox{\citep{park2012task, runtime_variability_lotz, rra_nico, temoto}}, and there are few works that fully address TACA~\citep{braberman2018extended, task_co_adapt_camara_garlan}. 
More critically, to the best of our knowledge, the existing solutions for TACA  require a significant and complex re-programming of the adaptation logic for each different use case, including the creation of multiple models based on different DSLs\mbox{~\citep{task_co_adapt_camara_garlan}} or implementing the managing subsystem itself~\citep{braberman2018extended}, hindering the adoption of SAS methods in robotics.

To address the limitations of SAS methods for TACA, this paper proposes to extend traditional robotics architectures with a novel knowledge-based managing subsystem for RObot Self-Adaptation~(\acronym) that promotes reusability, composability, and extensibility. The main novelty of \acronym is its knowledge base~(KB) which captures 
knowledge about the actions the robot can perform, the robot's architecture, the relationship between both, and their requirements to answer questions such as ``What actions can the robot perform in situation X?'' and  ``What is the best configuration available for each action in situation Y?'', for example, ``Can the robot perform an inspection action when the battery level is lower than 50\%?'' or ``What is the best software configuration for the inspection action when the visibility is low?''. 
This results in a reusable solution for TACA in which all application-specific aspects of the adaptation logic are captured in its KB.

In addition to a conceptual framework, this work provides a reference implementation of \acronym as an open-source framework that can be reused for research on self-adaptive robotic systems.  \acronym is implemented as a ROS~2-based system~\cite{ros2}, leveraging \emph{TypeDB}~\citep{dorn2023type, typeql} for knowledge representation and reasoning, and \emph{behavior trees}~(BT)~\citep{colledanchise2018behavior} as well as \emph{PDDL-based planners}~\citep{pddl} for task decision-making.

The \emph{feasibility} of using \acronym for runtime self-adaptation in robotic systems is demonstrated by applying it to the SUAVE exemplar~\citep{suave}, and its adaptation \emph{performance} is evaluated in comparison to other approaches available in the exemplar. \acronym's \emph{reusability} is demonstrated by using it to model the TACA scenarios described by \cite{braberman2018extended} and \cite{task_co_adapt_camara_garlan}. 
The \emph{development effort} for using \acronym is evaluated by analyzing the number of elements contained in the knowledge models created to solve the aforementioned use cases and comparing it with the size of a BT-based approach used to solve SUAVE. \acronym's \emph{development effort scalability} is demonstrated by showing how \acronym's knowledge model grows with the addition of extra adaptations in a hypothetical scenario.

In summary, the main contributions of this paper are:
\begin{enumerate}
\item a \emph{modular architecture} for self-adaptive robotic systems that extends robotics architectures with a managing subsystem and supports reusability, composability, and extensibility;
\item a \emph{reusable knowledge model} to capture all application-specific aspects of the adaptation logic required for TACA in self-adaptive robotic systems;
\item a reference \emph{open-source implementation} of the framework that can be reused for self-adaptive systems research; and
\item an \emph{experimental evaluation} of \acronym based on simulated robotic self-adaptation scenarios.
\end{enumerate}

The remainder of this paper is organized as follows. \Cref{sec:example} describes the TACA use case used to exemplify and evaluate this work. \Cref{sec:related_works} presents related works. \Cref{sec:architecture} describes how this work proposes to extend robotics architectures with \acronym. \Cref{sec:kb} details \acronym's KB. \Cref{sec:realization} describes the proposed reference implementation of \acronym. \Cref{sec:evaluation} showcases \acronym's evaluation. \Cref{sec:conclusion} concludes this work and presents future research directions.

\section{Running example}\label{sec:example}
Throughout this paper, the SUAVE exemplar~\citep{suave} is used as an example to ease the understanding of the proposed solution, and later, it is used to evaluate \acronym.

SUAVE consists of an Autonomous Underwater Vehicle (AUV) used for underwater pipeline inspection. The AUV's mission consists of performing the following actions in sequence: \emph{(A1) searching for the pipeline} and \emph{(A2) simultaneously following and inspecting the pipeline}. When performing its mission, the AUV is subject to two uncertainties: \emph{(U1) thruster failures}, and \emph{(U2) changes in water visibility}. These uncertainties are triggers for parameter and structural adaptation. When \emph{U1} occurs while performing \emph{A1} or \emph{A2}, the AUV activates a functionality to recover its thrusters.
When \emph{U2} happens while performing \emph{A1}, the AUV adapts its search altitude.

To demonstrate TACA, this work extends SUAVE with an \emph{(A3) recharge battery} action and a \emph{(U3) battery level} uncertainty. With these extensions, the AUV's battery level can suddenly drop to a critical level, requiring the AUV to abort the action it is performing \emph{(A1 or A2)} and perform \emph{A3}. In this situation, the AUV needs to perform TACA by adapting its task execution and architecture to perform \emph{A3}.
To better evaluate \acronym by serving as a baseline for comparison, this work extends the SUAVE exemplar with a managing subsystem where the adaptation logic is implemented with BTs, and the AUV's architectural variants as well as the architectural adaptation execution are realized with System Modes~\citep{system_modes}\footnote{The SUAVE exemplar is already configured to use System Modes}. Furthermore, this work introduces a new \emph{reaction time} metric that represents the time a managing system takes to react to uncertainties and adapt the managed subsystem.

\section{Related work}\label{sec:related_works}

This work combines principles from self-adaptive systems and knowledge representation and reasoning to design a reusable framework for developing adaptive robotics architectures. \Cref{sec:related_rob_archs} analyzes existing robotics architectures and describes the architectural patterns from robotics architectures adopted in this work. \Cref{sec:related_sars} reviews related research on self-adaptive robotic systems that leverage knowledge representation techniques to promote reusability, as well as studies that consider the relationship between task execution and architectural adaptation. Additionally, it discusses how these works influenced the design of the proposed framework and highlights its distinctions from existing approaches.

\subsection{Robotics Architectures}\label{sec:related_rob_archs}

Numerous approaches have been proposed for programming and designing autonomous robot architectures~\mbox{\citep{Kortenkamp2016}}. In recent years, two main trends have emerged: component-based frameworks and middlewares--among which ROS~\cite{ros2} stands out due to its widespread adoption in academia and industry--and layered architectures \mbox{\citep{barnett2022architectural}}. \mbox{\cite{barnett2022architectural}} reviewed 21 robotics architectures and concluded that most architectures follow a layered pattern, and even those that do not can still have their elements mapped onto a layered architectural structure. Furthermore, they found that all architectures include a bottom functional layer responsible for interacting with the robot's hardware, an upper task decision layer--whose responsibilities vary across architectures--and an arbitrary number of intermediate layers. This work aims to design a reusable solution for TACA that can be integrated into robotics architectures adhering to these architectural patterns. To achieve this, the proposed solution establishes a clear separation between architectural management and task logic, organizing them into distinct layers, or subsystems, as commonly referred to in the self-adaptive systems community.

The LAAS architecture~\mbox{\citep{laas_98}} is an example of a three-layered architecture consisting of a functional layer, an executive, and a decision layer. The functional layer contains the robot's control and perception algorithms. The executive layer receives a task plan from the decision layer and selects functions from the functional layer to realize each action in the task plan. The decision layer includes a planner that generates task plans and a supervisor responsible for monitoring plan execution and triggering replanning when necessary. 
More recent examples of layered robot architectures include AEROSTACK~\mbox{\citep{Sanchez-Lopez-2016}}, designed for aerial drone swarms, and SERA~\mbox{\citep{Garcia-2018}}, which is tailored for decentralized and collaborative robots. These architectures build on the layered model but focus on providing domain-specific solutions.

Cognitive architectures~\mbox{\citep{Kotseruba-2018}}, such as CRAM~\mbox{\citep{Beetz-2010, Kazhoyan-2021}}, focus on generating intelligent and flexible behavior by integrating cognitive capabilities such as planning, perception, or reasoning. However, being integral solutions, these works provide a blueprint for the complete robot control system and are not intended for reuse and integration with other methods, thus making it difficult to adapt and customize for specific applications.

\subsection{Self-adaptive robotic systems}\label{sec:related_sars}

Despite advances in self-adaptive robotic systems, fully addressing task and architectural co-adaptation (TACA) with reusable and scalable solutions remains an open challenge. While some studies explore the relationship between task execution and architectural adaptation, only a few explicitly address TACA—and those that do face limitations in reusability and practical applicability in robotics. This work aims to bridge this gap by introducing a knowledge-based framework that can capture the necessary knowledge to solve TACA across different use cases, can be directly applied to robotic systems, and supports modular modifications for incorporating different adaptation strategies.

\mbox{\cite{ALBERTS2025112258}} recently conducted a systematic mapping study\footnote{\mbox{\cite{ALBERTS2025112258}} do not claim that the mapping study is a complete overview of the literature but rather a characterization of the field} on ``robotics software architecture-based self-adaptive systems''~(RSASSs), identifying 37 primary studies on RSASSs published since 2011. Among these, \cite{ALBERTS2025112258} identified that four studies~\citep{park2012task, runtime_variability_lotz, rra_nico, task_co_adapt_camara_garlan} consider, to varying degrees, the relationship between the tasks a robot performs and architectural adaptation. A non-systematic snowballing of the primary studies identified by \cite{ALBERTS2025112258} revealed two additional studies~\citep{braberman2018extended, temoto} that also explore this relationship.

In the context of knowledge-based methods, \cite{ALBERTS2025112258} identified six studies~\citep{park2012task, stefan_adaptive_runtime_models, graph_based_nico, p20_stefan_ice, p26_darko_jasper, suave} that use knowledge representation techniques to capture knowledge required for the adaptation logic. Among these, \citep{p26_darko_jasper, suave} do not propose solutions for RSASSs but instead demonstrate the application of Metacontrol~\citep{hernandez2018self, bozhinoski2022mros} in different robotics use cases.
While these methods do not address TACA, they provide valuable insights for designing knowledge-based approaches to self-adaptation.

\begin{table}[]
\centering
\caption{Related frameworks for robot self-adaptation. Where A-t-T means ``Architectural state triggers task adaptation'', and T-t-A means ``Task triggers architectural adaptation''.}
\label{tab:related_works}
\resizebox{\textwidth}{!}{%
\begin{tabular}{lllllllll}
\hline
\multicolumn{1}{c}{\multirow{2}{*}{\textbf{Approach}}} &
  \multicolumn{2}{c}{\textbf{\begin{tabular}[c]{@{}c@{}}Architectural \\ adaptation\end{tabular}}} &
  \multicolumn{2}{c}{\textbf{TACA}} &
  \multicolumn{2}{c}{\textbf{Reusability}} &
  \multirow{2}{*}{\textbf{\begin{tabular}[c]{@{}l@{}}Robotics \\ Middleware\end{tabular}}} &
  \multirow{2}{*}{\textbf{\begin{tabular}[c]{@{}l@{}}Applied \\ to robot\end{tabular}}} \\ \cline{2-7}
\multicolumn{1}{c}{} &
  \textbf{Parameter} &
  \textbf{Structural} &
  \textbf{A-t-T} &
  \textbf{T-t-A} &
  \textbf{Conceptual} &
  \textbf{\begin{tabular}[c]{@{}l@{}}Software \\ Available\end{tabular}} &
   &
   \\ \hline
ICE~\citep{p20_stefan_ice}             & No           & Yes          & No           & No           & No           & No           & None           & No                 \\
\cite{graph_based_nico}                & No           & Yes          & No           & No           & No           & No           & None           & Real Robot         \\
Metacontrol~\citep{bozhinoski2022mros} & Yes          & Yes          & No           & No           & Yes          & Yes          & ROS 1\&2       & Real Robot         \\
SHAGE~\citep{park2012task}             & No           & Yes          & No           & Yes          & Partially    & No           & None           & No                 \\
\cite{runtime_variability_lotz}        & No           & Yes          & No           & Yes          & No           & No           & None           & No                 \\
RRA~\citep{rra_nico}                   & Yes          & Yes          & No           & Partially    & Yes          & Yes          & ROS 1          & Simulated          \\
TeMoto~\citep{temoto}                  & Yes          & Yes          & No           & Yes          & Partially    & Yes          & ROS 1          & Real robot         \\
MORPH~\citep{braberman2018extended}    & Yes          & Yes          & Yes          & Yes          & No           & No           & None           & No                 \\
\cite{task_co_adapt_camara_garlan}     & Yes          & Yes          & Yes          & Yes          & Partially    & No           & ROS 1          & Simulated          \\
\textbf{ROSA}                          & \textbf{Yes} & \textbf{Yes} & \textbf{Yes} & \textbf{Yes} & \textbf{Yes} & \textbf{Yes} & \textbf{ROS 2} & \textbf{Simulated} \\ \hline
\end{tabular}%
}
\end{table}

\cite{park2012task} introduced the SHAGE framework for task-based and resource-aware architecture adaptation in robotic systems. SHAGE partially solves TACA, as it can adapt the robot's architecture at runtime to specifically realize each action in its task plan when it needs to be performed. However, SHAGE does not support task execution adaptation based on the robot's architectural state. 
SHAGE promotes reusability by leveraging architectural models and knowledge captured with an ontology to reason about adaptation at runtime. However, to the best of our knowledge, there is no implementation of the SHAGE framework that works with common robotic frameworks. Thus, it is not possible to directly reuse SHAGE.

\mbox{\cite{runtime_variability_lotz}} proposed a method to model operational and quality variability using two distinct domain-specific languages~(DSLs) models. Their work provides a high-level discussion on how these models could be used at runtime to enable architectural adaptation based on the actions executed by the robot. An interesting aspect of their approach is the clear separation between functional and non-functional requirements: one model captures the task deliberation logic, functional requirements, and their variation points, while the other focuses on non-functional requirements and their possible variations.
While this separation of concerns simplifies the modeling process, combining task deliberation with functional requirements reduces reusability. Any change in the task deliberation logic directly impacts the modeling of functional requirements, making the approach less flexible. Additionally, they do not provide sufficient details on how these models are used at runtime, nor do they present an evaluation to demonstrate the feasibility of their approach.

\mbox{\cite{rra_nico}} proposed RRA as a model-based approach for structural, parameter, and connection adaptation in robotic systems. Their method employs six distinct models to capture all the knowledge required for adaptation. These models represent the robot's architecture and its variability, its functionalities and their variability, the mapping between functions and architecture, the required interfaces (i.e., inputs, outputs, and data types) for the adaptation logic, and the adaptation logic itself. RRA  models the dependencies between the tasks a robot can perform and the architectural configurations needed to accomplish them. This is achieved by decomposing each task into multiple functionalities and capturing the available architectural variants for realizing each function. At deployment time, the robot's operator selects a task, and RRA manages only the functionalities required for that task. Although RRA considers the relationship between tasks and architecture to some extent, it cannot be classified as TACA, as this dependence is only accounted for at deployment time. Architectural adaptation occurs based on the selected task rather than dynamically at runtime in response to the individual actions the robot needs to perform. Moreover, RRA does not adapt the task execution based on the robot's architectural state.

\cite{temoto} proposed the TeMoto as a general solution for robotic systems' dynamic task and resource management. TeMoto partially solves TACA, as it can adapt the robot's architecture to realize the actions being performed by the robot. TeMoto does not completely fulfill TACA as it cannot adapt the task execution given the robot's architectural state. TeMoto provides reusable mechanisms for resource management, but reusability is limited since the adaptation logic must be implemented for all managed resources, and the knowledge about the dependencies between actions and architecture are programmatically included in the actions' code.

\cite{braberman2018extended} proposed MORPH as a reference architecture to enable TACA. They showcased on a conceptual level how MORPH can be applied to enable TACA in an unmanned aerial vehicle~(UAV) use case. However, since MORPH is only demonstrated at a conceptual level, it is hard to evaluate the feasibility of applying MORPH to robotic systems at runtime. To the best of our knowledge, there is no framework implementing the complete MORPH architecture. Therefore, it is not possible to directly reuse MORPH.

In the context of TACA, \cite{task_co_adapt_camara_garlan} developed a method for finding \emph{optimal} task and reconfiguration plans for an autonomous ground vehicle~(AGV) navigating in a graph-like environment. To enable optimal planning within reasonable time limits, their method first reduces the search space by finding all possible reconfiguration plans and then computing the shortest N paths the robot can take to reach its goal. Then, it uses this information along with task-specific models that capture mission quality attributes (e.g., energy consumption, collision probabilities) and a preferred utility to apply model checking and determine an optimal reconfiguration plan for each path. Finally, an optimization function selects the best plan based on a predefined utility function (e.g., minimizing energy consumption, time, or collision probability). Although their approach reduces the planning search space to improve planning time, their experiments show that solving the navigation use case still takes an average of $15.1$ seconds, an impractical duration for robots that frequently need to replan at runtime to handle uncertainties. Additionally, while the approach is model-based, it relies on task-specific model transformations (e.g., converting the map or battery model into PRISM model snippets), which require dedicated implementations for different tasks. This limits the reusability of their approach for different types of tasks, as it requires a considerable amount of development effort.

\cite{graph_based_nico} argue that robots should have access to and exploit software-related knowledge about how they were engineered to support runtime adaptation. They demonstrate how labeled property graphs~(LPGs) can be used to persistently store and compose different domain models specified with domain-specific languages~(DSL) to enable runtime architectural adaptation. Although their approach is interesting, its reusability is limited as it does not define a knowledge model that can be reused for other applications: for each different use case and DSL the roboticist is responsible for creating a translation from the DSL to the corresponding LPG. 

\mbox{\cite{stefan_adaptive_runtime_models, p20_stefan_ice}} propose ICE as a method for adapting the information processing subsystem in multi-robot systems, specifically by adapting the connections between system components. Their approach uses an ontology to define each component's required inputs and outputs, along with quality-of-service information for each connection. This ontology is then translated into answer set programming (ASP), and an ASP solver determines an optimal configuration. While they conceptually demonstrate how their method could be applied to robots, they do not demonstrate it with a robotic system. Moreover, their approach focuses solely on structural and connection adaptation to maintain the functionality of the information processing subsystem, without considering other subsystems of the robotic system.

\mbox{\cite{hernandez2018self, bozhinoski2022mros}} proposed Metacontrol as a knowledge-based solution for parameter and structural adaptation. Metacontrol leverages the TOMASys~\citep{hernandez2018self} ontology to capture the knowledge required for the adaptation logic and to reason at runtime to decide when and how the system should adapt. Similar to RRA, Metacontrol decomposes the system into functionalities, using TOMASys to represent the robot's functionalities, the architectural variants that implement each functionality, and the non-functional requirements associated with these variants. However, Metacontrol cannot perform TACA as TOMASys does not capture the relationship between the system functionalities and the robot's actions.

In conclusion, existing works that fully address TACA~\mbox{\citep{braberman2018extended, task_co_adapt_camara_garlan}} face limitations in reusability and practical applicability in robotics. The authors of MORPH~\mbox{\citep{braberman2018extended}} note that no complete system has been developed, and they have not demonstrated its feasibility. While the approach proposed by \mbox{\cite{task_co_adapt_camara_garlan}} provides the most complete solution to TACA in the literature, it also faces reusability limitations by relying on multiple distinct domain-specific language (DSL) models and requiring task-specific implementations for model transformations. To address these limitations, this work proposes capturing all the knowledge required for adaptation logic in a single knowledge model. This approach requires only one model--conforming to the proposed knowledge model--to be designed for each application. Additionally, an open-source reference implementation of ROSA is provided, enabling researchers to reuse, extend, and build upon the proposed solution.

On the other hand, previous knowledge-based methods for RSASS\citep{ park2012task, stefan_adaptive_runtime_models, p20_stefan_ice, graph_based_nico, hernandez2018self, bozhinoski2022mros} do not capture all knowledge required to enable TACA. They either lack the ability to capture the relationship between the robot's actions and architecture \citep{graph_based_nico, hernandez2018self, bozhinoski2022mros}, or the knowledge required to decide when and how the robotic system should adapt~\mbox{\citep{park2012task}}. Although these works do not capture all the knowledge required for TACA, they are able to capture, to varying degrees, knowledge that supports RSASSs. Thus, this work takes inspiration from them to design ROSA's knowledge model while addressing their limitations. More concretely, ROSA's knowledge model is designed to capture the relationship between the robot's actions and architecture and the knowledge required to decide when and how TACA should be performed.

\section{Architecture}\label{sec:architecture}
\begin{figure}[h]
    \centering
    \includegraphics[width=0.5\linewidth]{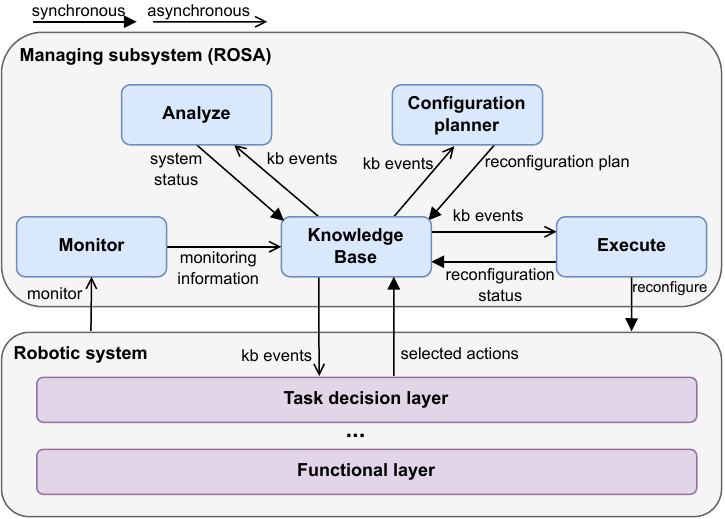}
    \caption{The upper layer depicts \acronym's architecture and the bottom layer depicts the robotic system.}
    \label{fig:architecture}
\end{figure}

To enable TACA, this paper proposes to extend traditional robotics architectures with \acronym, using it as a managing subsystem for the robotic subsystem~(see \Cref{fig:architecture}). This section first describes the assumptions made about the robotics architecture and the requirements to use it alongside \acronym, and then it details \acronym's architecture.

\subsection{Robotics architecture}
This work assumes that the robotics architecture is layered, containing a bottom functional layer, an upper task decision layer, and an arbitrary number of layers in between, as common in robotics architectures~\citep{barnett2022architectural}. The functional layer is responsible for interacting with the robots' sensors and actuators, and the task decision layer is responsible for task planning and execution\footnote{Some architecture have distinct layers for handling task planning and execution, in the context of this work, they can be considered as sub-layers of the task decision layer.}. To enable TACA with \acronym, the task decision layer shall use the knowledge contained in \acronym's KB to decide which actions to perform, and it must update the KB with the actions selected to be performed to enable \acronym to configure the robot's architecture accordingly. To enable architectural adaptation, the robotic architecture must be component-based, its components must be able to be activated and deactivated at runtime, and its components' parameters must be able to be adapted at runtime.  

\subsection{\acronym architecture}
\acronym's architecture adheres to the MAPE-K loop~\citep{kephart2003vision}. It \emph{monitors} the managed subsystem, \emph{analyzes} whether adaptation is required, when needed, \emph{plans} how the managed subsystem should be reconfigured, and \emph{executes} the selected reconfigurations. All these steps interact with a central \emph{KB}.

To promote reusability, composability, and extensibility, the architecture is designed with the following premises: \emph{(1)} \emph{all} knowledge required for the adaptation logic is captured in the central KB, \emph{(2)} there is no inter-component communication between the MAPE components~\citep{weyns_patterns_2013}, \emph{(3)} the MAPE components insert and read data from or to the KB via standardized interfaces, and \emph{(4)} there is no explicit coordination between the MAPE-K components. Premise \emph{1} promotes reusability by only requiring the modeling of the relevant knowledge for applying \acronym to different applications in a single model. Premises \emph{1} to \emph{4} promote composability and extensibility by allowing the MAPE components to be stateless and self-contained.
 
\section{Knowledge base}\label{sec:kb}
To fulfill the architectural premise that all knowledge required for the adaption logic should be captured in a central KB, this work proposes a KB component composed of the knowledge model depicted in  \Cref{fig:k_model} and the set of rules depicted in \Cref{fig:status_diagrams}, described in \Cref{sec:kb_model} and \Cref{sec:rules} respectively.

The knowledge model is presented as a conceptual data model~(CDM) conforming to a particular case of the enhanced entity-relationship (EER)\mbox{\cite{herm, thalheim2013entity}} model, an extension of the entity-relationship model \mbox{~\cite{chen1976entity}} that accounts for subclassing and higher-order relationships, i.e., relations between relationships. The EER model captures information as entities, relationships, and attributes. An entity is a ``'thing' which can be distinctly identified''\mbox{~\cite{chen1976entity}}, a relationship is an association among entities or relationships, and an attribute represents a property of an entity or relationship\mbox{~\cite{herm, thalheim2013entity}}. In addition, entities and relationships play a role in the relationships they are part of, which is identified by the label on the arrows in \mbox{\Cref{fig:k_model}}. This CDM was selected since it supports n-ary relationships, many-to-many relationships, higher-order relationships, and attributes for both entities and relationships. \mbox{\Cref{sec:rep_req} details why these representation capabilities are relevant to the proposed model}. The rules are also described at a conceptual level as decision diagrams. \mbox{\Cref{sec:impl}} presents the details on how the knowledge model and rules can be implemented and executed and runtime.

\subsection{Knowledge model}\label{sec:kb_model}
To enable adaptation, the knowledge model captures
\emph{what} can be adapted with the architectural knowledge depicted in \Cref{fig:arch_model}; \emph{why} to adapt and \emph{how} to select an adaptation with the adaptation heuristics knowledge depicted in \Cref{fig:adapt_model}; and \emph{how} to execute an adaptation with the reconfiguration plan knowledge depicted in \Cref{fig:reconfig_plan}. Tables \ref{tab:elements_km_architectural}-\ref{tab:elements_km_reconfig} define each element in the model alongside examples based on SUAVE to ease its understanding.  

\setcounter{figure}{2}
\setcounter{subfigure}{0}
\begin{subfigure}
\setcounter{figure}{2}
\setcounter{subfigure}{0}
    \centering
    \begin{minipage}[b]{0.123\textwidth}
        \centering
        \includegraphics[width=\linewidth]{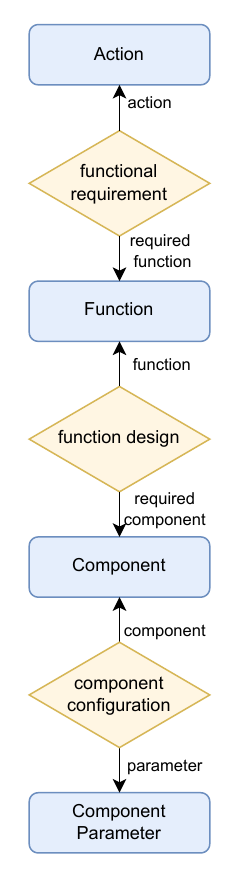}
        \caption{Architectural}
        \label{fig:arch_model}
    \end{minipage}  
    \hfill
\setcounter{figure}{2}
\setcounter{subfigure}{1}
    \centering
    \begin{minipage}[b]{0.482\textwidth}
        \centering
        \includegraphics[width=\linewidth]{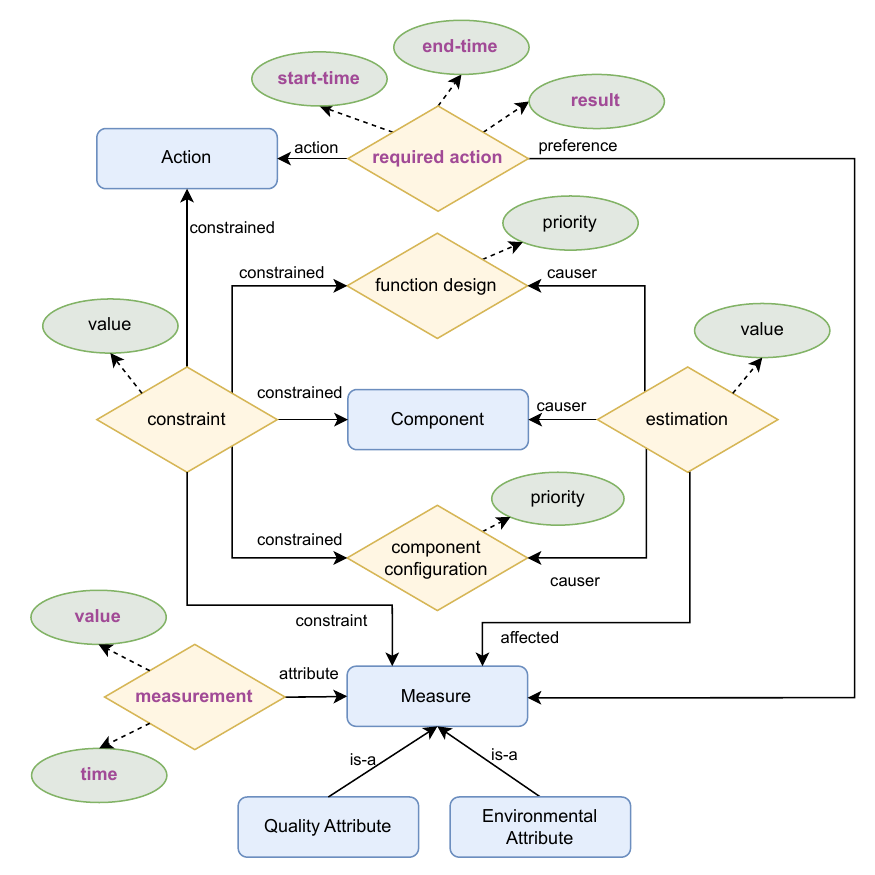}
        \caption{Adaptation heuristic}
        \label{fig:adapt_model}
    \end{minipage}
    \hfill
\setcounter{figure}{2}
\setcounter{subfigure}{2}
    \centering
    \begin{minipage}[b]{0.354\textwidth}
        \centering
        \includegraphics[width=\linewidth]{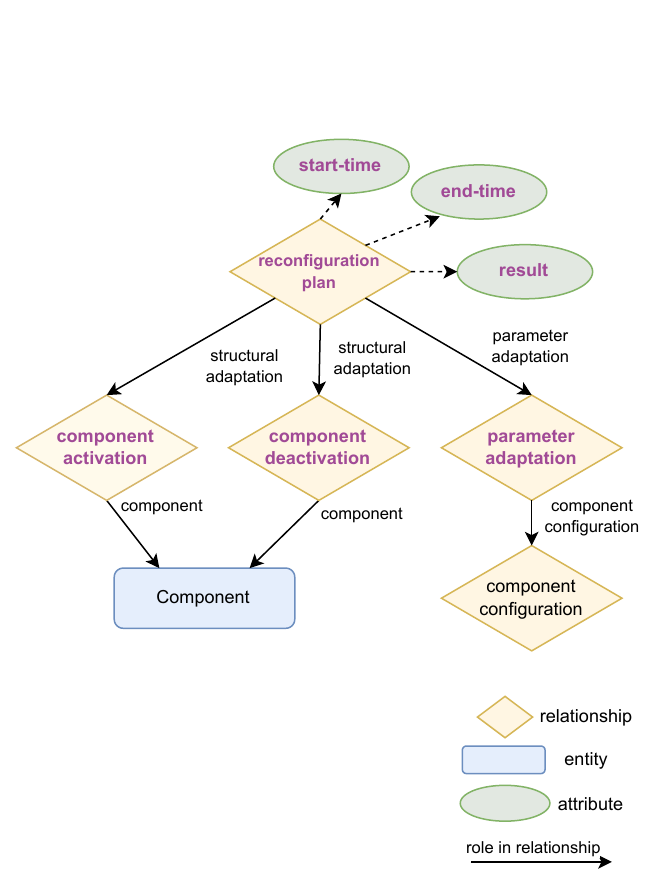}
        \caption{Reconfiguration plan}
        \label{fig:reconfig_plan}
    \end{minipage}

\setcounter{figure}{2}
\setcounter{subfigure}{-1}
    \caption{\acronym's knowledge model. Architectural knowledge is on the left. Adaptation heuristic knowledge in the center. Reconfiguration Plan knowledge on the right. The labels on the arrows represent which role an entity or relationship plays in a relationship. Instances of the entities, relationships, and attributes with \runtimefont are created at runtime. Instances of the other entities, relationships, and attributes are defined at design time.}
    \label{fig:k_model}
\end{subfigure}

\subsubsection{Architectural knowledge}
The architectural knowledge~(see \Cref{tab:elements_km_architectural}) captures what actions the robot can accomplish, the set of functionalities the robot needs to realize an action, the set of components required to realize a functionality, and the possible parameters for a component. 

\begin{table}[h]
\small  
\centering
\caption{The elements of the architectural knowledge. Instances of the elements with \runtimefont are created at runtime. Instances of the other elements are defined at design time. The `@key' expression after an attribute indicates that the attribute is used as a unique identifier. E stands for entity and R stands for relationship.}
\label{tab:elements_km_architectural}
\resizebox{\linewidth}{!}{%
\begin{tabular}{p{.01\linewidth}p{.12\linewidth}p{0.3\linewidth}p{0.31\linewidth}p{0.26\linewidth}}
\hline
\textbf{} &
  \textbf{Name} &
  \textbf{Definition} &
  \textbf{Example} &
  \textbf{Attributes} \\ \hline
\multirow{4}{*}{E} &
  Action & 
  ``Action is defined as an operation applied by an agent or team to affect a change in or maintain either an agent's state(s), the environment, or both.''\citep{ieep1872}. Where in the context of this paper the agent is the robot &
  An AUV can perform the \entity{Action} \instance{search pipeline}, and \instance{inspect pipeline} &
  \attribute{name:String @key}  \attribute{\runtime{status:String}} \attribute{\runtime{is-required:Bool}} \\
 &
  Function &
  ``A function is defined by the transformation of input flows to output flows''~\citep{wiki:sebok}, i.e., a function represents what the system can do &
  An AUV can have several \entity{Functions}, like \instance{generate search path} and \instance{generate path to follow pipeline}. The \instance{generate search path} function can be considered a transformation of the AUV's current position (input) to a goal waypoint (output) &
  \attribute{name:String  @key} \attribute{always-improve:Bool} \attribute{\runtime{status:String}} \attribute{\runtime{is-required:Bool}} \\
 &
  Component &
  Hardware and software parts that compose the system &
  A \instance{thruster} is a hardware component, and a \instance{generate spiral search path node} is a software component &
  \attribute{name:String @key} \attribute{always-improve:Bool} \attribute{\runtime{status:String}} \attribute{\runtime{is-required:Bool}} \attribute{\runtime{is-active:Bool}} \attribute{\runtime{pid:Integer}} \\
 &
  Component \hspace{1cm} Parameter &
  A specific parameter configuration of a \entity{Component} &
  The \instance{generate spiral search path node} component can be configured with different search altitudes. Each search altitude is represented as a distinct \entity{Component Parameter} &
  \attribute{key:String} \attribute{value:String} \\ \hline
\multirow{3}{*}{R} &
  funcional \hspace{1cm} requirement &
  Represents which \entity{Functions} a certain \entity{Action} requires to be performed &
  The \instance{search pipeline} action requires the \instance{control motion}, \instance{maintain motion}, \instance{localization}, \instance{detect pipeline}, \instance{generate search path}, and \instance{coordinate mission} functions &
  N/A \\
 &
  function design  &
  Represents a solution for a \entity{Function} as a set of \entity{Components} \citep{hernandez2018self} &
  The \instance{generate search path} function can have multiple \relationship{function designs}, e.g., one using the \instance{generate spiral search path node} component, and another one using the \instance{generate lawnmower search path node} component &
  \attribute{name:String @key} \attribute{priority:Integer} \attribute{\runtime{status:String}} \attribute{\runtime{is-selected:Bool}} \\
 &
  component configuration &
  Represents a possible configuration of a \entity{Component} as a set of \entity{Component Parameters} &
  The \instance{generate spiral search path node} component can be configured in different ways by combining different values of its search altitude and speed parameters. Each combination is represented as an instance of a \relationship{component configuration} &
  \attribute{name:String @key} \attribute{priority:Integer} \attribute{\runtime{status:String}} \\ \hline
\end{tabular}%
}
\end{table}

The architectural knowledge enables parameter adaptation by capturing each parameter configuration of a component with a distinct \relationship{component configuration} relationship, relating one \entity{Component} to a set of \entity{Component Parameters}. It enables structural adaptation by capturing the different possibilities for solving a system functionality as distinct \relationship{function design} relationships which relate a \entity{Function} to a set of \entity{Components}. It enables TACA by indirectly capturing the dependencies between the robot's actions and architecture with the \relationship{functional requirement} relationship which relates an \entity{Action} to the \entity{Functions} it requires.

At runtime, \acronym performs parameter adaptation by switching the selected \relationship{component configurations}. It performs structural adaptation by changing the selected \relationship{function designs} and consequently the active \entity{Components}. The task decision layer in combination with \acronym performs TACA by selecting suitable \relationship{function designs} and \relationship{component configurations} for each \entity{Action} the robot needs to perform, and with the task decision layer selecting different \entity{Actions} to perform according to the feasible configurations.

\subsubsection{Adaptation heuristic knowledge}
This work considers that the robotic system might need to adapt due to changes in the environment, changes in the system's quality attributes (QAs)~\citep{wiki:sebok}, component failures, and changes in the robot's selected actions.

\begin{table}[h]
\small  
\centering
\caption{The elements of the adaptation heuristic knowledge. Instances of the elements with \runtimefont are created at runtime. Instances of the other elements are defined at design time. The `@key' expression after an attribute indicates that the attribute is used as a unique identifier. E stands for entity and R stands for relationship.}
\label{tab:elements_km_heuristic}
\resizebox{\linewidth}{!}{%
\begin{tabular}{p{.01\linewidth}p{.12\linewidth}p{0.3\linewidth}p{0.31\linewidth}p{0.26\linewidth}}
\hline
\textbf{} &
  \textbf{Name} &
  \textbf{Definition} &
  \textbf{Example} &
  \textbf{Attributes} \\ \hline
\multirow{3}{*}{E} &
  Measure &
  ``Measure is defined as a function over observations, state variables, and parameters.''\citep{ieep1872}
  &
  An AUV can have the \entity{Measures} \instance{battery level} or \instance{water visibility}&
  \attribute{name:String @key} \\
 &  Quality \hspace{1cm} Attribute &
  Quality Attribute is defined as a function over the system's state variables. Which, according to the Software Engineering Body of Knowledge~\citep{wiki:sebok}, can be considered as ``System functional and non-functional requirements used to evaluate the system performance'' &
  An AUV can have the \entity{Quality Attributes} \instance{battery level}, \instance{battery consumption}, and \instance{safety level} &
  \attribute{name:String @key} \\
 &
  Environmental Attribute &
  Environmental Attribute is defined as a function over observations of the environment. That is, it represents a metric of the environment with respect to a certain attribute &
  An underwater environment can have \instance{water visibility} as an \entity{Environmental Attribute} &
  \attribute{name:String @key} \\ \hline
\multirow{4}{*}{R} &
  \runtime{required action} &
  Represents the \entity{Actions} required at runtime &
  The \instance{inspect pipeline} \entity{Action} can be required to be performed at runtime, leading \acronym to configure the AUV appropriately to carry out the inspection action &
  \attribute{\runtime{start-time:Datetime}} \attribute{\runtime{end-time:Datetime}} \attribute{\runtime{result:String}} \\
 &
  \runtime{measurement} &
  ``Measurement is defined as the act of evaluating the measures''~\citep{ieep1872} &
  The \instance{battery level} \entity{Quality Attributes} can have a measurement of 0.5 &
  \attribute{\runtime{value:Double}} \attribute{\runtime{time:Datetime}} \\
 &
  constraint &
  Represents \entity{Measure} constraints for performing a \entity{Action} or for selecting a \relationship{function design}, \entity{Component}, or \relationship{component configuration} &
  If the \instance{water visibility} environmental attribute is low, the \instance{high altitude} configuration for the \instance{generate spiral search path node} cannot be selected &
  \attribute{operator:String} \attribute{value:Double} \attribute{\runtime{status:String}}  \\
 &
  estimation &
  Represents the estimated impact of a \relationship{function design}, \entity{Component}, or \relationship{component configuration} on a \entity{Measure} &
  A lamp \entity{Component} is expected to positively impact the \instance{water visibility} when turned on &
  \attribute{value:Double} \attribute{type:String} \\ \hline
\end{tabular}%
}
\end{table}

The adaptation heuristic knowledge enables adaptation due to changes in the environment or the system's QAs with the \relationship{constraint} relationship by capturing constraints on the selection of \entity{Actions}, \entity{Components}, \relationship{function designs}, or \relationship{component configurations} in terms of measured values of \entity{Measures}. 
It enables adaptation due to component failures by capturing a \entity{Component}'s status as an attribute and adaptation due to changes in the robot's task execution by capturing what \entity{Actions} need to be performed with the \relationship{required action} relationship. At runtime, adaptation is triggered when a measurement violates a constraint, when a component has a failure status, or when required actions change. 

To capture the decision criteria on how to select an adaptation, the \attribute{priority} attribute can be used to express the order of priority for selecting each \relationship{function design} or \relationship{component configuration}. For more complex criteria, the \relationship{estimation} relationship can be used to capture the estimated impact of selecting a \relationship{function design}, \entity{Component}, or \relationship{component configuration} on the measured values of \entity{Measures}. Furthermore, a \relationship{required action} can relate to a \entity{Measure} to indicate the preferred \relationship{estimation} when selecting a configuration for that action. At runtime, the configuration planner component exploits this knowledge to decide which configuration to select.

\subsubsection{Reconfiguration plan knowledge}
The reconfiguration plan knowledge represents which component parameters must be updated (i.e., how to execute parameter adaptation) and which components must be activated and deactivated (i.e., how to execute structural adaptation). The execute component exploits this knowledge at runtime to reconfigure the managed subsystem.

\begin{table}[h]
\small  
\centering
\caption{The elements of the reconfiguration plan knowledge. Instances of the elements with \runtimefont are created at runtime. Instances of the other elements are defined at design time. R stands for relationship.}
\label{tab:elements_km_reconfig}
\resizebox{\linewidth}{!}{%
\begin{tabular}{p{.01\linewidth}p{.12\linewidth}p{0.3\linewidth}p{0.31\linewidth}p{0.26\linewidth}}
\hline
\textbf{} &
  \textbf{Name} &
  \textbf{Definition} &
  \textbf{Example} &
  \textbf{Attributes} \\ \hline
\multirow{4}{*}{R} &
  \runtime{reconfiguration plan} &
  Represents the reconfiguration plan generated in the plan step & When the AUV completes the \instance{search pipeline} action and starts the \instance{inspect pipeline} action, the \relationship{reconfiguration plan} consists of deactivating the \instance{generate spiral search path node} and activating the \instance{follow pipeline node}
   & 
  \attribute{\runtime{start-time:Datetime}} \attribute{\runtime{end-time:Datetime}} \attribute{\runtime{result:String}} \\
 &
  \runtime{component activation} &
  Represents the components that should be activated & When the AUV starts the \instance{inspect pipeline} action, the \relationship{component activation} relates to the \instance{follow pipeline node}
   & N/A
   \\
 &
  \runtime{component deactivation} &
  Represents the components that should be deactivated & When the AUV completes the \instance{search pipeline} action, the \relationship{component deactivation} relates to the \instance{generate spiral search path node}
   & N/A
   \\
 &
  \runtime{parameter adaptation} &
  Represents the component parameters that should be updated &
  When the water visibility changes, the \relationship{parameter adaptation} consists of a parameter adaptation of the component configuration of the \instance{generate spiral search path node} & N/A
   \\ \hline
\end{tabular}%
}
\end{table}

\subsection{Rules}\label{sec:rules}
To enable \acronym to reason about when the managed subsystem should be adapted and what type of adaptation is needed, the KB contains a set of generic rules that define how the status of the system can be inferred based on monitored information, e.g., how to infer if a constraint is violated and how to propagate a constraint violation. 
A change in status only occurs when measurements are updated or when a component fails. The complete status inference rules are depicted as decision diagrams in \Cref{fig:status_diagrams}. 

The status of the required \entity{Components}, \entity{Functions}, and \entity{Actions} indicate when and what type of adaptation is needed. When a \entity{Component} is ``unsolved'' or in ``configuration error''~(see \Cref{fig:component_status}), parameter adaptation is required, i.e., a new \relationship{component configuration} needs to be selected. When a required \entity{Function} is ``unsolved'' or in ``configuration error''~(see \Cref{fig:function_status}), structural adaptation is required, i.e., a new \relationship{function design} needs to be selected. When a required \entity{Action} is ``unfeasible''~(see \Cref{fig:action_status}), TACA is needed, i.e., the task decision layer must choose a new action to perform, consequently triggering architectural adaptation.

\setcounter{figure}{3}
\setcounter{subfigure}{0}
\begin{subfigure}
\setcounter{figure}{3}
\setcounter{subfigure}{0}
    \centering
    \begin{minipage}[b]{0.32\textwidth}
        \includegraphics[width=\linewidth]{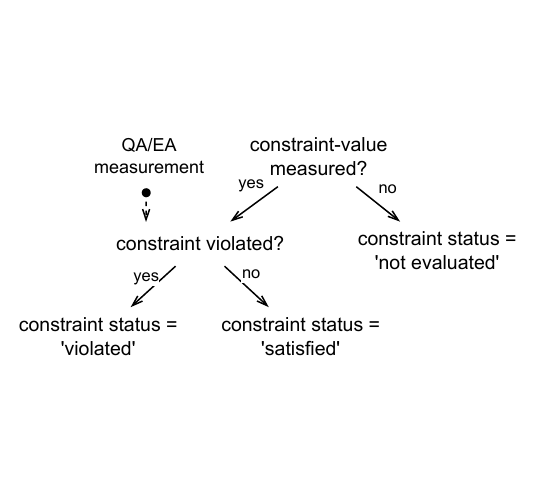}
        \caption{Constraint status}
        \label{fig:constraint_status}
    \end{minipage}  
   \hfill
\setcounter{figure}{3}
\setcounter{subfigure}{1}
    \begin{minipage}[b]{0.32\textwidth}
       \includegraphics[width=\linewidth]{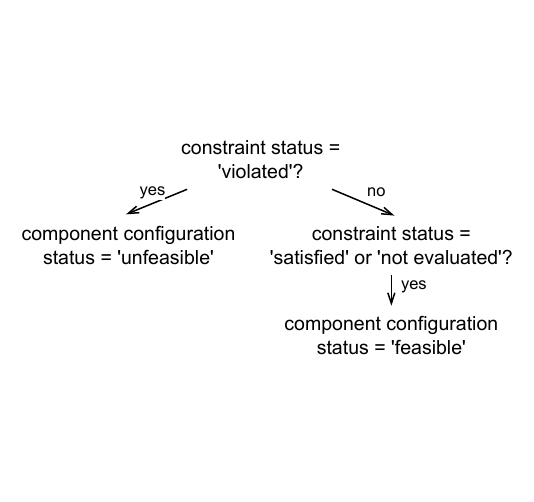}
        \caption{Component configuration status}
        \label{fig:component_config_status}
    \end{minipage}
    \hfill
\setcounter{figure}{3}
\setcounter{subfigure}{2}
    \begin{minipage}[b]{0.32\textwidth}
       \includegraphics[width=\linewidth]{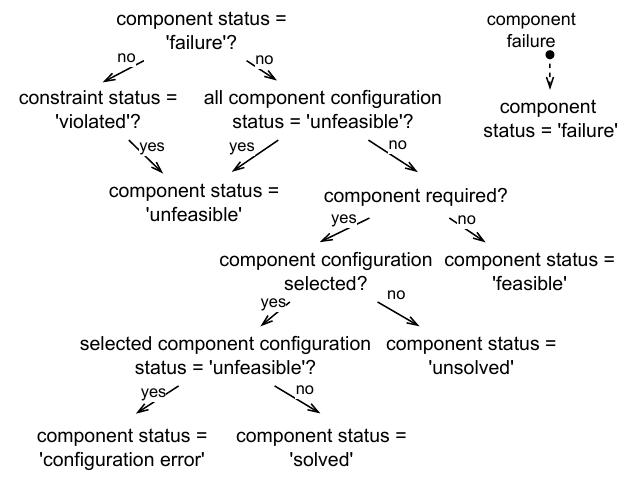}
        \caption{Component status}
        \label{fig:component_status}
    \end{minipage}
    \hfill
\setcounter{figure}{3}
\setcounter{subfigure}{3}
    \begin{minipage}[b]{0.32\textwidth}
      \includegraphics[width=\linewidth]{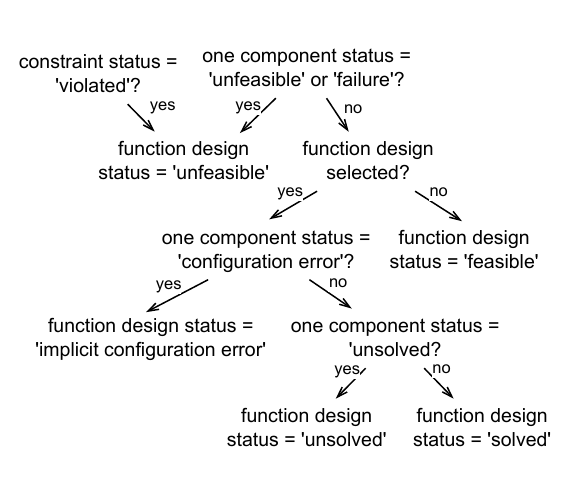}
        \caption{Function design status}
        \label{fig:function_design_status}
    \end{minipage}
    \hfill
\setcounter{figure}{3}
\setcounter{subfigure}{4}
    \begin{minipage}[b]{0.32\textwidth}
       \includegraphics[width=\linewidth]{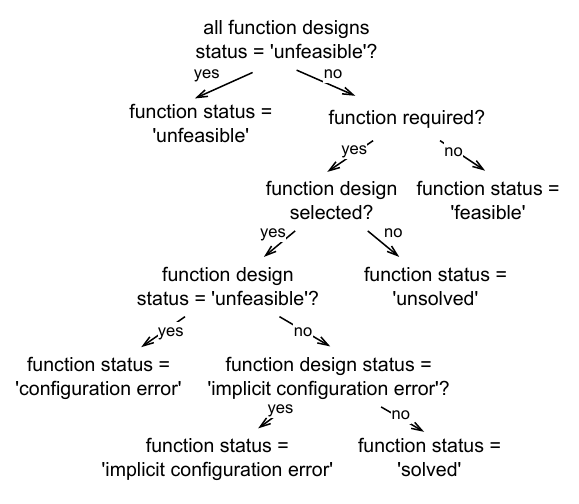}
        \caption{Function status}
        \label{fig:function_status}
    \end{minipage}
    \hfill
\setcounter{figure}{3}
\setcounter{subfigure}{5}
    \begin{minipage}[b]{0.32\textwidth}
       \includegraphics[width=\linewidth]{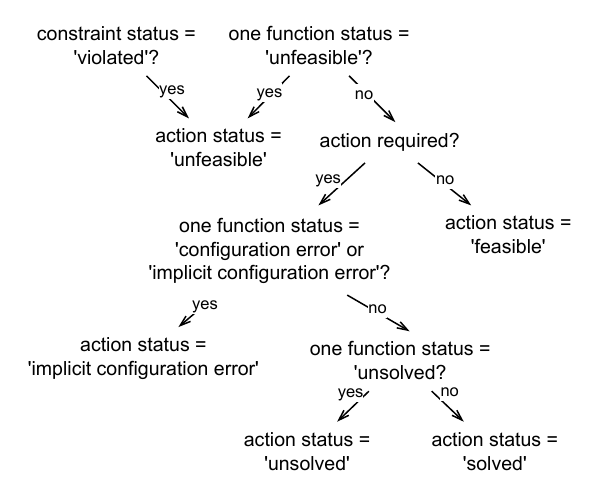}
        \caption{Action status}
        \label{fig:action_status}
    \end{minipage}

\setcounter{figure}{3}
\setcounter{subfigure}{-1}
    \caption{Rules used to infer the status of the elements in the knowledge model, presented as decision diagrams}
    \label{fig:status_diagrams}
\end{subfigure}

\section{Architecture realization}\label{sec:realization}
This section details the proposed reference implementation for \acronym. First, the representation requirements imposed by the proposed knowledge model are analyzed to select a suitable knowledge representation technique. Afterward, the proposed implementation is explained.

\subsection{Representation requirements}\label{sec:rep_req}
The proposed knowledge model was presented as a conceptual data model~(CDM) in a non-machine-readable format. To transform the knowledge model into a machine-readable format while maintaining its semantics and structure, its representation requirements must be identified and used to select a suitable technology to implement it. The representation requirements imposed by the proposed knowledge model are:
\begin{enumerate}
    \item n-ary relationships: the knowledge model contains relationships with different arity\footnote{the number of distinct elements that can be part of a relationship}, e.g., the \relationship{constraint} relationship has arity~5, and the \relationship{measurement} relationship has arity~1;
    \item many-to-many relationships: the knowledge model contains relationships with different cardinalities\footnote{the number of element instances that can be part of a relationship}, e.g., \relationship{function design} is a 1~\entity{Function}-to-many~\entity{Components} relationship;
    \item \label{enum:3} higher-order relationships: some relationships relate relationships to relationships, e.g., the \relationship{constraint} relationship;
    \item \label{enum:4} attributes for entities and relationships: both entities and relationships have attributes, e.g., the \entity{Action} entity and the \relationship{constraint} relationship.
\end{enumerate}

These requirements limit which technology can be used. For example, technologies using graph-based or descriptive logic-based knowledge representation techniques (e.g., OWL~\citep{Antoniou2004}) do not satisfy the representation requirements above apart from allowing attributes for entities (part of Requirement~\ref{enum:4}). Although it is possible to transform the proposed knowledge model into one that can be represented with graph-based or description logic-based approaches by reifying the model~\citep{olive2007conceptual}, each reification applied to the original model can be considered as an introduction of a semantic disparity between the CDM and the machine-readable model, making it harder to understand and reuse it. Thus, this work does not consider applying reification.

A technology that satisfies all representation requirements is TypeDB~\citep{dorn2023type, typeql}. TypeDB is a polymorphic database based on type theory that implements the polymorphic entity-relation-attribute~(PERA)~\cite{typeql} data model. The PERA model subsumes the CDM used as the meta-model for the proposed ROSA model, allowing it to be implemented without modifications. Furthermore, TypeDB has a reasoning system that is able to reason over rules of the form $ antecedent \Rightarrow consequent$ to infer new facts at query time. Where $antecedent$ represents a precondition for inferring the $consequent$ and is expressed as a first-order logic expression combining elements from the model (i.e., entities, relationships, and individuals), and the $consequent$ is a single new fact inferred when the $antecedent$ holds true. The ROSA model rules presented in \Cref{fig:status_diagrams} can be implemented with TypeDB without modifications. For these reasons, this work uses TypeDB to implement the proposed knowledge model and rules.

\subsection{Implementation}\label{sec:impl}

\acronym is implemented as a ROS 2-based system, where the MAPE-K components~(depicted in \Cref{fig:architecture}) are realized as ROS nodes, and interfaces are implemented using ROS services or topics. The proposed \acronym implementation uses ROS~(Robot Operating System) as its robotics framework since ROS is the current de facto standard robotics framework, and it has been designed, among other things, to promote software reusability in the robotics ecosystem~\cite{ros2}. In this implementation, ROS handles the communication between system components, schedules callbacks for incoming messages and events, and manages the lifecycle of ROS nodes.
The full \acronym implementation is available at \url{https://github.com/kas-lab/rosa}\footnote{In addition to ROSA, this work provides a generic ROS package to integrate ROS~2 with TypeDB: \url{https://github.com/kas-lab/ros_typedb}}.

\subsubsection{Knowledge base and analyze}
The KB component consists of the TypeDB implementation of the proposed knowledge model and inference rules, the ROS interfaces for communicating with the MAPE components, and the logic to manage \acronym's knowledge which is stored in a TypeDB database. In this reference \acronym implementation, TypeDB's reasoner fulfills the role of the analyze component, executing \acronym's inference rules~(\Cref{fig:status_diagrams} to infer new data when the KB is queried. Thus, since TypeDB's reasoner is part of TypeDB, there is no separate analyze component. 

\textit{Knowledge model and rules:} 
To exemplify how the knowledge model is implemented, \Cref{lst:km} depicts how the \relationship{functional-requirement} relationship and the \entity{Action} entity are defined with TypeQL (TypeDB's query language). Line 1 defines the \relationship{functional-requirement} relationship, and lines 2-3 define that it can relate elements that play the role of \instance{actions} and \instance{required-functions}. Lines 5-7 define the \entity{Action} entity and that it has the attributes `action-name' (its unique identifier) and `action-status'. Lines 8-9 define that it can play the \instance{action} role in a \relationship{functional-requirement} relationship and the \instance{constrained} role in a \relationship{constraint} relationship. 

\begin{lstlisting}[caption=TypeQL query to define \relationship{functional-requirement} and \entity{Action}, label=lst:km, columns=fullflexible, language=TypeQL]
functional-requirement sub relation,
  relates action,
  relates required-function;

Action sub entity,
  owns action-name @key,
  owns action-status,
  plays functional-requirement:action,
  plays constraint:constrained;
\end{lstlisting}

\Cref{lst:rules} exemplifies how the inference rules are implemented with TypeQL. The \entity{component-status-configuration-error} rule defines that a \entity{Component} has a `configuration error' status when it is required, it does not have an `unfeasible' or `failure' status, and it is in a \relationship{component-configuration} relationship that is selected and has an `unfeasible' status~(see \Cref{fig:component_status}). 

\begin{lstlisting}[caption=TypeQL rule to infer whether a component is in `configuration error', label=lst:rules, columns=fullflexible, language=TypeQL]
rule component-status-configuration-error:
 when {
  $c isa Component, has is-required true;
  not {
    $c has status $c_status; 
    $c_status like 'unfeasible|failure'; 
  };
  (component: $c) isa component-configuration,
    has is-selected true,  
    has status 'unfeasible';
 } then {
   $c has status 'configuration error'; 
 };
\end{lstlisting}

\textit{Interfaces:} 
The KB component abstracts the details of interacting with TypeDB with the ROS interfaces it implements, enabling the MAPE components to read and write knowledge via the interfaces described in \Cref{tab:ros_interfaces}. 
When the MAPE components request or send data to the KB component via these interfaces, the KB component queries the TypeDB database to retrieve or write knowledge. For example, when the task decision layer calls the service \emph{/action/selectable} to retrieve the name of the selectable \entity{Actions} (i.e., actions that do not have an `unfeasible' status), the KB component performs the TypeQL query depicted in \Cref{lst:typedb_query} to retrieve the name (unique identifiers) of the selectable \entity{Actions}. 
When data is written in the KB, the KB component publishes a message in the \emph{/events} topic specifying which type of data was written, i.e., `monitoring data', `action update', `reconfiguration plan'. Additionally, the KB component provides the \emph{/query} service, which can be used to perform \emph{any} TypeDB query to the database. It is not used in ROSA's runtime workflow, but it enables users to perform custom queries, for example, to retrieve all reconfiguration plans that were executed.

\begin{table}[h]
\centering
\caption{ROS interfaces}
\label{tab:ros_interfaces}
\begin{tabular}{lll}
\hline
\multicolumn{3}{l}{\textbf{Topics}} \\ \hline
\textbf{Publisher} &
  \textbf{Subscriber} &
  \textbf{Name} \\ \hline
KB &
  \begin{tabular}[c]{@{}l@{}}Configuration Planner\\ Execute\\ Task Decision Layer\end{tabular} &
  $\sim$/events \\ \hline
Monitor nodes &
  KB &
  /diagnostics \\ \hline
\multicolumn{3}{l}{\textbf{Services}} \\ \hline
\textbf{Server} &
  \textbf{Client} &
  \textbf{Name} \\ \hline
\multirow{4}{*}{KB} &
  Configuration Planner &
  \begin{tabular}[c]{@{}l@{}}$\sim$/function/adaptable\\ $\sim$/function\_designs/selectable\\ $\sim$/function\_designs/priority\\ $\sim$/component/adaptable\\ $\sim$/component\_configuration/selectable\\ $\sim$/component\_configuration/priority\\ $\sim$/select\_configuration\end{tabular} \\ \cline{2-3} 
 &
  Execute &
  \begin{tabular}[c]{@{}l@{}}$\sim$/reconfiguration\_plan/get\_latest\\ $\sim$/reconfiguration\_plan/result/set\\ $\sim$/component/active/set\\ $\sim$/component\_parameters/get\end{tabular} \\ \cline{2-3} 
 &
  Task Decision Layer &
  \begin{tabular}[c]{@{}l@{}}$\sim$/action/selectable\\ $\sim$/action/request\end{tabular} \\ \cline{2-3} 
 &
  User &
  $\sim$/query \\ \hline
\end{tabular}
\end{table}

\begin{figure}
\begin{lstlisting}[caption=TypeDB query to fetch selectable actions' names, label=lst:typedb_query, columns=fullflexible, language=TypeQL]
match  $a isa Action, has name $name;
  not {$a has status 'unfeasible';};
  fetch $name;
\end{lstlisting}
\end{figure}

\subsubsection{Monitor} 
\acronym's implementation does not provide generic monitor nodes. They should be implemented as needed for each application with the requirement that they publish the monitored information in the \emph{/diagnostics} topic\footnote{The \emph{/diagnostics} topic is a standard topic for publishing system diagnosis information within the ROS ecosystem. See \href{https://www.ros.org/reps/rep-0107.html}{ROS REP 107} for more information.} with the standard ROS \emph{DiagnosticArray} message format. 
When a monitor node sends measurement updates to the KB, the message field in the \emph{DiagnosticStatus} message needs to be set to `QA measurement' or `EA measurement', and when sending component status updates (e.g., that the component is in failure), the message field must be set to `Component status'. When the KB receives monitoring data, it sends an event message in the `/events' topic to inform that monitoring data was written in the KB.

\subsubsection{Configuration planner} 
The configuration planner component selects the configurations (i.e., \relationship{function designs} or \relationship{component configuration}) with the highest priority. When the configuration planner receives an event message indicating that monitoring data was written in the KB or that there was an update in the required actions, it calls the services `/function/adaptable' and `/component/adaptable' to check which \entity{Functions} and \entity{Components} must be adapted. Then, it calls the services `/function/selectable' and `/component/selectable' to check which \relationship{function designs} and \relationship{component configurations} are available for the \entity{Functions} and \entity{Components} that need to be adapted. Finally, the configuration planner selects the \relationship{function designs} and \relationship{component configurations} with the highest priority and informs the KB about the newly selected configuration by calling the service `/select\_configuration'. When this service is called, the KB component checks the current state of the robot, creates a \relationship{reconfiguration plan} to bring the robot to the goal configuration, and sends an event message in the `/events' topic to inform that there is a new reconfiguration plan available.

\subsubsection{Execute}  
\begin{figure}[h]
    \centering
    \includegraphics[width=.4\linewidth]{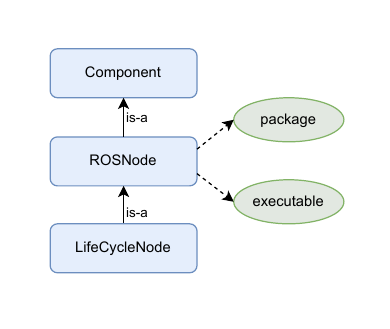}
    \caption{ROS specific knowledge}
    \label{fig:ros_schema}
\end{figure}

In ROS-based systems, software components are realized either as ROS nodes or as a particular type of ROS nodes called \emph{lifecycle} nodes. The difference between both is that the latter can be set to different states at runtime, such as \emph{active} and \emph{inactive}, and the former cannot. To enable ROSA to leverage ROS~2 mechanisms to adapt the system, the knowledge model was extended to capture knowledge about ROS~2 components as depicted in \Cref{fig:ros_schema}. The execute component performs structural adaptation by starting or killing ROS nodes or switching the state of lifecycle nodes to active or inactive, and it performs parameter adaptation by calling the ROS's parameter API to change the ROS nodes' parameters at runtime.

When the execute component receives an event message indicating that a new reconfiguration plan was added to the KB, it calls the service `/reconfiguration\_plan/get\_latest' to get the latest \relationship{reconfiguration plan}. Then, it adapts the robot's architecture according to the reconfiguration plan. Finally, it calls the services `/reconfiguration\_plan/result/set' and `/component/active/set' to update the KB with the result of the reconfiguration plan and which components are active.

\subsubsection{Task decision layer}
\acronym's implementation provides an integration for the task decision layer for both PDDL-based planners and BTs, which are implemented leveraging the PlanSys2~\citep{plansys2} and the BehaviorTree.CPP\footnote{\url{https://www.behaviortree.dev/}} packages, respectively.

\emph{Planning:} 
To enable task decision-making and execution with PDDL-based planners in combination with \acronym, the planner and plan executor must consider the runtime feasibility of performing the robot's actions as inferred by the KB component. 
This work maps the action status from \acronym's knowledge model to PDDL by 
capturing whether the action's status is feasible as a PDDL predicate of the form ``\emph{action\_feasible ?action}'' and using it as a precondition to select the respective action. An example can be seen in~\Cref{lst:task_interface_pddl} where the action \emph{my\_action} can only be selected when it does not have an `unfeasible' status in the KB.
At runtime, if an action becomes unfeasible during execution, the plan executor triggers re-planning to generate a new action plan. This results in task execution adaptation and, if the newly selected actions require a different architectural configuration, also in architectural adaptation, i.e., TACA.

\begin{lstlisting}[caption=PDDL formulation example for ROSA, label=lst:task_interface_pddl, columns=fullflexible, language=PDDL]
 (:durative-action my_action
   :parameters (?a - action ...)
   :duration (...)
   :condition (and
     (over all (my_action_action ?a))
     (over all (action_feasible ?a))
     ...
   )
   :effect (and ...)
 )
\end{lstlisting}

To handle the interaction between PlanSys2 and \acronym's KB, this work provides a custom ROS~2 node called \emph{RosaPlanner} and a custom PlanSys2 action called \emph{RosaAction}. The \emph{RosaPlanner} is responsible for querying the KB and updating the PDDL problem formulation with information on whether the ROSA actions are feasible or not using the aforementioned ``\emph{action\_feasible ?action}'' PDDL predicate. The \emph{RosaAction} action is responsible 
for querying the KB to request or cancel an \entity{Action} when the execution of an action starts or finishes. Each PlanSys2 action that should be managed by \acronym should derive from \emph{RosaAction}, and it should implement the logic for the specific action execution. 

\begin{figure}[h]
    \centering
    \includegraphics[width=0.5\linewidth]{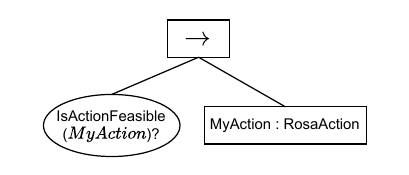}
    \caption{BT pattern for TACA with \acronym. The \emph{IsActionFeasible} condition node takes the action name as a parameter. The \emph{MyAction} node derives from the proposed \emph{RosaAction} node and implements the action execution. The action names in the BT must match the names defined in the KB.}
    \label{fig:tasK_interface_bt}
\end{figure}

\emph{Behavior trees:} 
To enable task decision-making and execution with BTs in combination with \acronym, the BTs must consider the runtime feasibility of performing the robot's actions as inferred by the KB component. This work proposes adding before action nodes a condition node that queries the KB to ask whether the following action is feasible. The proposed pattern is depicted in \Cref{fig:tasK_interface_bt}, where the action \emph{MyAction} would only be executed when its status in the KB is not `unfeasible'.

To enable the use of the BehaviorTree.CPP package to implement BTs for \acronym and abstract away the interactions with the KB, this work implements a reusable custom condition node called \emph{IsActionFeasible} and a custom action node called \emph{RosaAction}. The condition node queries the KB to check whether an \entity{Action} is feasible before selecting it to be executed, and the action node queries the KB to request or cancel an \entity{Action} when the execution of an action starts or finishes, respectively.

\section{Evaluation}\label{sec:evaluation}

This section evaluates \acronym to answer the following questions:
\begin{itemize}
    \item \emph{Feasibility}: Is it feasible to use \acronym to enable runtime TACA in ROS 2-based robotic systems?
    \item \emph{Performance}: How does \acronym perform compared to other managing subsystems for ROS 2-based systems?
    \item \emph{Reusability}:  To what extent can \acronym's knowledge model capture the knowledge required for TACA?
    \item \emph{Development effort}: What is the development effort of using \acronym for adding adaptation to different robotic systems and how does it compare to other approaches?
    \item \emph{Development effort scalability}: How does the development effort of using \acronym for adding adaptation to robotic systems scale for more complex systems?
\end{itemize}

\subsection{Experimental design}

\subsubsection{Feasibility} To evaluate the feasibility of applying ROSA at runtime to enable TACA in ROS~2-based robotic systems, it was applied to the SUAVE exemplar described in \Cref{sec:example}. SUAVE was selected since, to the best of our knowledge, it is the only ROS~2-based open-source exemplar for self-adaptive robotic systems.

\subsubsection{Performance} 
To evaluate \acronym's performance, metrics were collected 
with the SUAVE exemplar.
The metrics collected were the ones available in SUAVE, \emph{search time} and \emph{distance of the pipeline inspected}, in addition to the \emph{reaction time} metric introduced in this paper. For the original use case, the experiments were performed with no managing subsystem, with a BT managing system, with Metacontrol\footnote{Metacontrol is the only managing subsystem packaged with SUAVE. The Metacontrol implementation for this use case is described in detail in the SUAVE paper~\citep{suave}.}, and with \acronym. For the extended use case, the experiments were performed with a BT managing system and with \acronym.

\subsubsection{Reusability} 
To evaluate to what extent ROSA's knowledge model can capture the knowledge required for TACA, it was used to model the TACA scenarios presented by ~\cite{task_co_adapt_camara_garlan} and ~\cite{braberman2018extended}\footnote{To the best of our knowledge, these are the only two scenarios in the literature in which the authors explicitly claim the need for TACA.}, and we showcase how the captured knowledge can be exploited to enable TACA. The experimental setup for both scenarios is not publicly available. Thus, it was not possible to apply \acronym at runtime to the simulated environments they used. 

\subsubsection{Development effort} To evaluate the development effort of using \acronym to enable adaptation in robotic systems, we analyze the number of elements contained in the \acronym model created to solve the use cases described in this paper. Furthermore, we compare it to the number of elements modeled with the BT approach to solve the SUAVE use case. 

\subsubsection{Development effort scalability} To analyze how \acronym's development effort scales for more complex use cases, we showcase how many elements must be modeled in \acronym's model to include structural and parameter adaptation in a hypothetical adaptation scenario.

\medskip

The experimental setup for the \emph{Feasibility} and \emph{Performance} evaluation is open source and reproducible, it can be found at 
\url{https://github.com/kas-lab/suave_rosa}. 
The models designed to evaluate \emph{Reusability} can be found at
\url{https://github.com/kas-lab/rosa_examples}.

\subsection{Feasibility}
\subsubsection{Experimental setup}
\begin{figure*}[h]
    \centering
    \includegraphics[width=.85\linewidth]{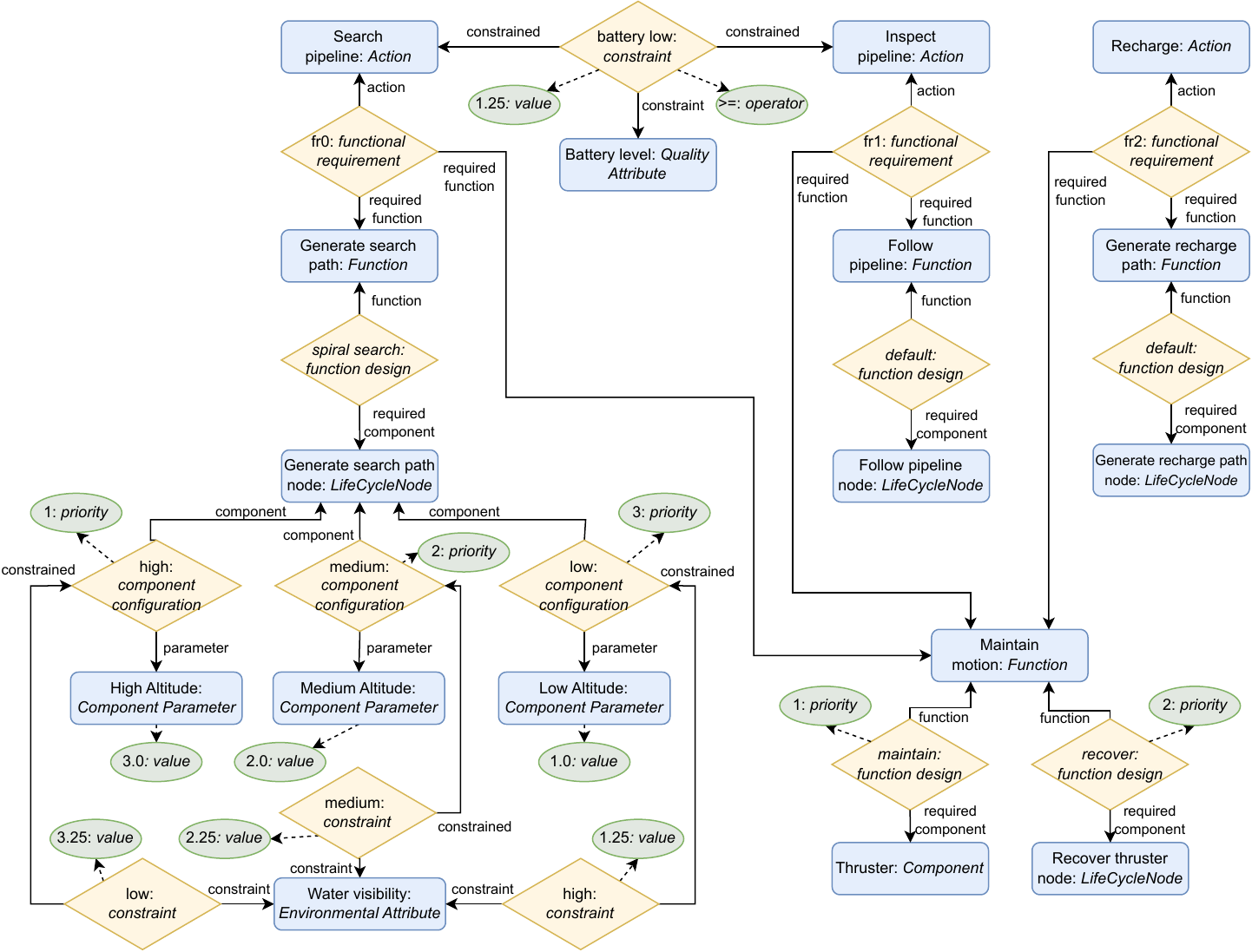}
    \caption{ROSA model for SUAVE.}
    \label{fig:suave_rosa_model}
\end{figure*}

To solve the adaptation scenarios of the SUAVE exemplar with \acronym, the model depicted in \Cref{fig:suave_rosa_model} was created.

\textit{Thruster failure}~($U_1$) was solved with structural adaptation by including two possible \relationship{function designs} of the \instance{maintain motion} function.
\textit{Runtime behavior:} when a thruster fails, the \instance{maintain} function design status is set to unfeasible (see \Cref{fig:function_design_status}), and it cannot be selected anymore. Then, the \instance{recover} function design is selected, and the \instance{recover thruster node} component is activated. If all thrusters are recovered, the \instance{maintain} function design status becomes feasible and is selected again.

\textit{Changing water visibility}~($U_2$) was addressed with parameter adaptation by including three \relationship{component configurations} for the \instance{generate spiral node}, each representing a different altitude for searching for the pipeline. Furthermore, a \instance{water visibility} \relationship{constraint} was added to each configuration, representing the minimum water visibility in which the configuration can be used. \textit{Runtime behavior:} if the measured water visibility is higher than $3.25$, the component configuration \instance{High} is selected since it has priority number one. If the water visibility drops below $3.25$, its constraint status is set to violated (see \Cref{fig:constraint_status}), and, consequently, its status is set to unfeasible (see \Cref{fig:component_config_status}). Depending on the water visibility, the component configuration \instance{Medium} or \instance{Low} is then selected. If the water visibility increases again above $3.25$, the \instance{High} component configuration status becomes feasible and is selected.

\textit{Critical battery level}~($U_3$), occurring only in the extended SUAVE use case, was solved with TACA by extending the knowledge model with a \instance{recharge} action, and a \instance{battery level} \relationship{constraint} to the \instance{search pipeline} and \instance{inspect pipeline} actions, representing the minimum battery level at which they can be selected.
\textit{Runtime behavior:} when the battery level drops below $0.25$, the status of both the \instance{search pipeline} and \instance{inspect pipeline} actions is set to unfeasible, and the task decision layer cannot select them anymore. Therefore, the task decision layer selects the \instance{recharge} action which also triggers structural adaptation.

Note that extending the solution to solve SUAVE's extended version only required the inclusion of additional knowledge of the \instance{recharge} action and the \relationship{constraint} to the \instance{search pipeline} and \instance{inspect pipeline} actions. This demonstrates that an existing application modeled with \acronym can easily be extended to solve additional adaptation scenarios.

\begin{figure}[h]
    \centering
    \includegraphics[width=0.7\linewidth]{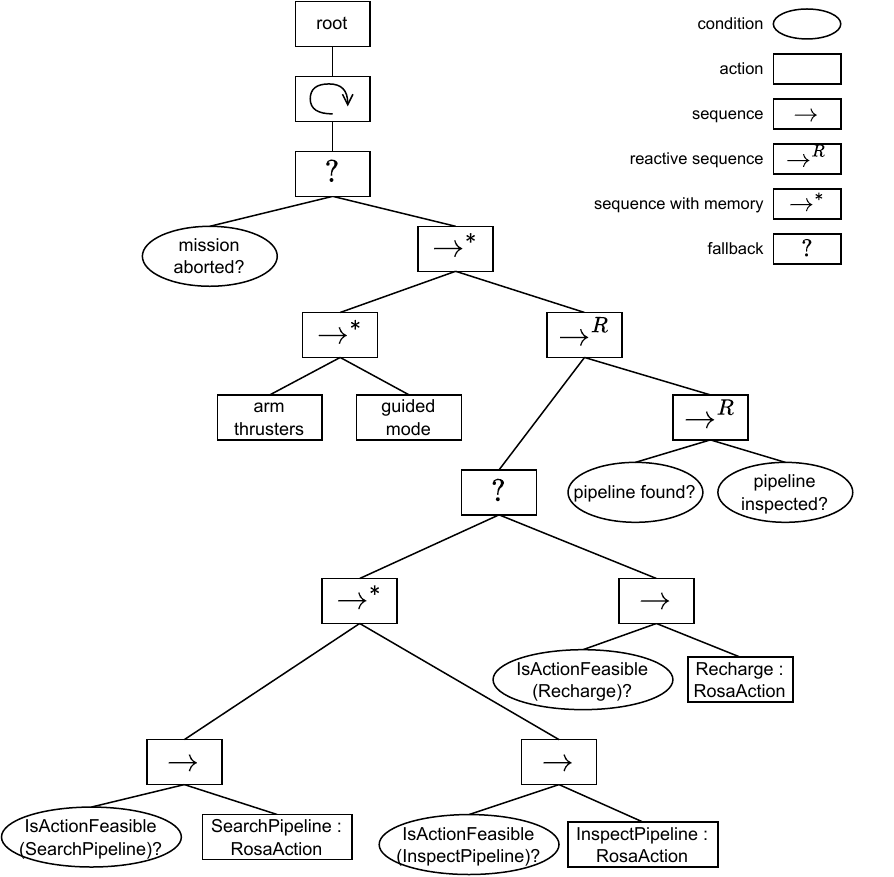}
    \caption{Behavior tree for the extended SUAVE use case}
    \label{fig:bt_extended}
\end{figure}

To apply \acronym in simulation to the SUAVE exemplar, the knowledge model depicted in \Cref{fig:suave_rosa_model} was implemented with TypeDB (see an example in \Cref{lst:suave}), and the AUV's mission was implemented with the BT depicted in \Cref{fig:bt_extended} as well with the PDDL formulation partially shown in \Cref{lst:suave_extended_pddl}.

\begin{lstlisting}[caption=Snippet of SUAVE's use case implementation in TypeDB, label=lst:suave, columns=fullflexible, language=TypeQL]
# Action search pipeline
$a_search_pipeline isa Action, has action-name
    'search_pipeline';

$f_generate_search_path isa Function, has function-name 
    'generate_search_path';

# functional-requirement relationship
(action: $a_search_pipeline, 
 required-function: $f_generate_search_path, 
 required-function: $f_maintain_motion) 
 isa functional-requirement;
\end{lstlisting}

\begin{figure}[h]
\begin{lstlisting}[caption=Snippet of the PDDL domain formulation for SUAVE containing the search pipeline action definition., label=lst:suave_extended_pddl, columns=fullflexible, language=PDDL]
 (:durative-action search_pipeline
   :parameters (?a - action ?p - pipeline ?r - robot)
   :condition (and
     (over all (robot_started ?r))
     (over all (search_pipeline_action ?a))
     (over all (action_feasible ?a))
   )
   :effect (and
     (at end(pipeline_found ?p))
   )
 )
 ...
\end{lstlisting}
\end{figure}

\subsubsection{Result} During the mission execution, the AUV was able to overcome all uncertainties: adapting to thruster failures ($U_1$) with structural adaptation, to changing water visibility ($U_2$) with parameter adaptation, and to an unexpected drop in the battery level ($U_3$) with TACA, demonstrating the feasibility of using \acronym to enable runtime TACA in ROS 2-based robotic systems.

\subsection{Performance}

\subsubsection{Experimental setup} To evaluate \acronym's performance, the SUAVE exemplar was configured with the same parameters as described in the SUAVE paper~\citep{suave}. In addition, for the extended use case, the battery was set to discharge within 200 seconds, and the \textit{search pipeline} and \textit{inspect pipeline} actions were set to require at least 25\% of battery to be performed.

\subsubsection{Results} 
The results obtained are shown in \Cref{tab:results}. 
The different managed subsystems had similar performance despite the difference in their reaction time. Furthermore, since the performance with \acronym was better than without a managing subsystem and close to the other 
 managing subsystems, it can be considered as additional evidence of the feasibility of applying it at runtime to enable self-adaptation in robotic systems.

\begin{table}[h]
\centering
\caption{Mission results}
\label{tab:results}
\resizebox{\linewidth}{!}{%
\begin{tabular}{p{.13\linewidth}p{.1\linewidth}p{.1\linewidth}p{.1\linewidth}p{.1\linewidth}p{.1\linewidth}p{.1\linewidth}p{.1\linewidth}p{.1\linewidth}}
\hline
\multirow{2}{*}{\begin{tabular}[c]{@{}l@{}}\textbf{Managing} \\ \textbf{system}\end{tabular}} &
  \multirow{2}{*}{\begin{tabular}[c]{@{}l@{}}\textbf{Number} \\ \textbf{of runs}\end{tabular}} &
  \multicolumn{2}{l}{\textbf{Search time (s)}} &
  \multicolumn{2}{l}{\textbf{Distance inspected (s)}} &
  \multicolumn{3}{l}{\textbf{Mean reaction time (s)}} \\ \cline{3-9} 
              &              & \textbf{Mean}          & \textbf{Std dev}       & \textbf{Mean}          & \textbf{Std dev}       & \textbf{U1}           & \textbf{U2} & \textbf{U3}        \\ \hline
\multicolumn{9}{l}{\textbf{SUAVE}}                                                                                             \\ \hline
None          & 100          & 174.75       & 36.00         & 33.20         & 13.49         & N/A          & N/A   & N/A         \\
BT            & 100          & 84.09          & 26.41          & 62.70          & 7.78          & 0.08          & 0.10   & N/A         \\
Metacontrol   & 100          & 89.24        & 35.57        & 60.57         & 11.17         & 1.55          & 0.82     & N/A       \\
\textbf{ROSA} & \textbf{100} & \textbf{85.11}  & \textbf{32.48} & \textbf{60.76} & \textbf{10.29} & \textbf{1.24} & \textbf{1.57} & \textbf{N/A} \\ \hline
\multicolumn{9}{l}{\textbf{SUAVE extended}}                                                                                    \\ \hline
BT            & 100          & 94.37          & 34.92          & 20.88          & 3.81          & 0.07           & 0.10   & 1.09          \\
\textbf{ROSA} & \textbf{100} & \textbf{92.75} & \textbf{35.92} & \textbf{18.97} & \textbf{3.38} & \textbf{1.39}  & \textbf{1.67} & \textbf{2.50}   \\ \hline
\end{tabular}%
}
\end{table}

\subsection{Reusability}\label{sec:reusability}
\subsubsection{Autonomous ground vehicle use case~\citep{task_co_adapt_camara_garlan}}
In this scenario, an AGV has to navigate from an initial to a goal position in a graph-like environment while facing uncertainties such as component failures, corridors with obstacles, and changing light conditions. 

\begin{figure}
    \centering
    \includegraphics[width=1\linewidth]{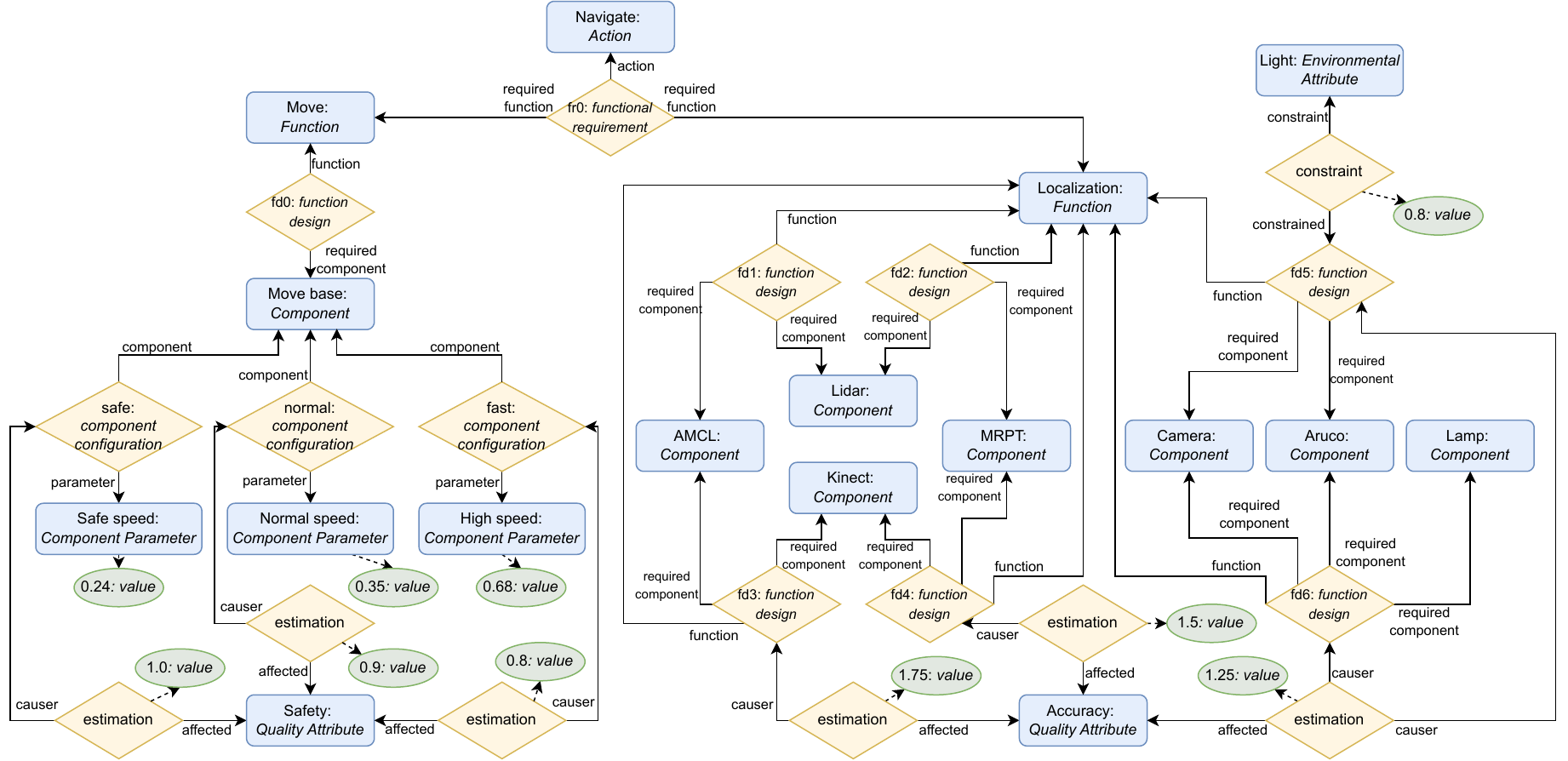}
    \caption{\acronym model for the AGV use case~\citep{task_co_adapt_camara_garlan}}
    \label{fig:camara}
\end{figure}

The AGV has distinct architectural variants available to solve navigation. It has three localization algorithms (AMCL, MRPT, or aruco) and three sensing components (camera, lidar, or Kinect). However, there are some restrictions on how they can be combined. The AMCL and MRPT algorithms can only be combined with lidar or Kinect, and the aruco algorithm can only be combined with a camera. In low-light conditions, the camera can only be used with a lamp. Furthermore, the robot can move at three different speeds. Each configuration has a different energy cost, safety, and accuracy level. This scenario can be solved with \acronym with the knowledge model depicted in \Cref{fig:camara}\footnote{The accuracy estimation for \emph{fd1} and \emph{fd2} and all energy estimations are omitted from the figure to improve readability}.

When navigating, the robot performs adaptation by selecting which corridors it needs to go through given its feasible configurations, and by selecting a suitable architecture configuration for each corridor it goes through. For example, to go from point $A$ to $B$, the AGV can go directly through a corridor with obstacles $C1$ or through corridors $C4\rightarrow C3\rightarrow C2$ without obstacles. Ideally, the AGV should go through $C1$ as it is the shortest path. Considering that the Kinect and AMCL combination is the only one with enough accuracy to go through a corridor with obstacles, in the case that the Kinect fails, the robot needs to perform TACA by adapting its task plan to go through $C4\rightarrow C3\rightarrow C2$ while simultaneously adapting its architecture, e.g., to use the lidar as its sensing component.

\subsubsection{Unmanned aerial vehicle use case~\citep{braberman2018extended}}\label{sec:morph_example}
In this scenario, a UAV has to search for samples in a predefined area and analyze them. To accomplish this mission, the UAV can perform the actions \emph{(A1) search for samples}, \emph{(A2) pick up and analyze samples}, \emph{(A3) analyze samples on site}, \emph{(A4) return to base and recharge}, and \emph{(A5) land and fold gripper}. The analyze action \emph{A2} performs a better analysis than \emph{A3}, however, it consumes more battery. Furthermore, action \emph{A2} requires a \emph{gripper}, while \emph{A3} requires an \emph{infra-red camera}. When operating, the UAV might run out of battery, and its gripper might fail.

There are three adaptation scenarios in this use case:
\emph{(1)} if the battery level is insufficient to perform \emph{A1}, the UAV must perform \emph{A4}; \emph{(2)} if the battery level is insufficient to perform \emph{A2} but it is still sufficient to perform \emph{A3}, the UAV must perform \emph{A3}; \emph{(3)} if the \emph{gripper} fails while performing \emph{A2}, the UAV must perform \emph{A3}. 
Before transitioning from \emph{A2} to \emph{A3}, the UAV must first perform \emph{A5}.

Adaptation scenarios \emph{1} and \emph{2} can be solved with \acronym's knowledge model by capturing the battery level as a \entity{Quality Attribute} and using it as a \relationship{constraint} for actions \emph{A1} and \emph{A2}. Adaptation scenario \emph{3} can be solved by capturing that the \entity{Function} required by \emph{A2} requires a \emph{gripper} component. Furthermore, the task decision layer is responsible for guaranteeing that \emph{A3} is only performed when \emph{A5} is finished.

\subsubsection{Results} This evaluation demonstrates that in addition to SUAVE, the knowledge required for the adaptation logic to solve the AGV and UAV use cases can be captured with \acronym's knowledge model. This indicates that \acronym's knowledge model can be used to capture the knowledge required for TACA in adaptation scenarios similar to the ones presented. 
Furthermore, it shows that all entities and relationships contained in the proposed knowledge model had to be used to model the aforementioned adaptation scenarios, supporting their inclusion in the knowledge model.

\subsection{Development effort}

\subsubsection{Experimental Setup} 
To measure \acronym's development effort for each use case, we count the number of entities and relationships contained in the models created for the use cases presented. To measure the BT managing system development effort for SUAVE,  we count the number of nodes contained in the BTs created to solve it, and the number of modes and parameters included in the System Modes' configuration file already packaged in SUAVE. In the remainder of this paper, we denote the elements modeled in both approaches as \emph{overhead}.

\subsubsection{Results}  The overhead for both approaches can be seen in \Cref{tab:dev_effort}. Although it is not possible to make a straightforward comparison between the development efforts of both approaches using the observed overheads since the difficulty of modeling an element using the different modeling techniques is subjective, analyzing the reason for the observed overheads provides insights for comparing the development efforts of both approaches.

\begin{table}[h]
\centering
\caption{Development effort of using ROSA and BTs as managing subsystems}
\label{tab:dev_effort}
\begin{tabularx}{\linewidth}{XXXX}
\hline
\multicolumn{4}{p{\linewidth}}{\textbf{ROSA}}                                                    \\ \hline
\textbf{Use case} & \textbf{Entities}      & \textbf{Relations} & \textbf{Total} \\ \hline
SUAVE             & 18                     & 12                     & 30             \\
SUAVE  extended    & 22                     & 16                     & 38             \\
AGV      & 18                     & 36                     & 54             \\
UAV               & 24                     & 16                     & 40             \\ \hline
\multicolumn{4}{p{\linewidth}}{\textbf{Behavior tree}}                                           \\ \hline
\textbf{Use case} & \textbf{BT} & \textbf{System modes}  & \textbf{Total} \\ \hline
SUAVE             & 27                     & 30                     & 57             \\
SUAVE extended    & 34                     & 38                     & 72             \\ \hline
\end{tabularx}%
\end{table}

\emph{\acronym:} 
In TypeDB, each entity and relationship is inserted in the KB with TypeQL queries such as the ones presented in \Cref{lst:suave}. 
Thus, the total number of queries that the roboticist must define to solve an adaptation scenario is equal to the sum of the number of entities and relationships contained in the model, with a clear separation between the task and adaptation logic. 

\begin{figure}[h]
    \centering
        \includegraphics[width=0.7\linewidth]{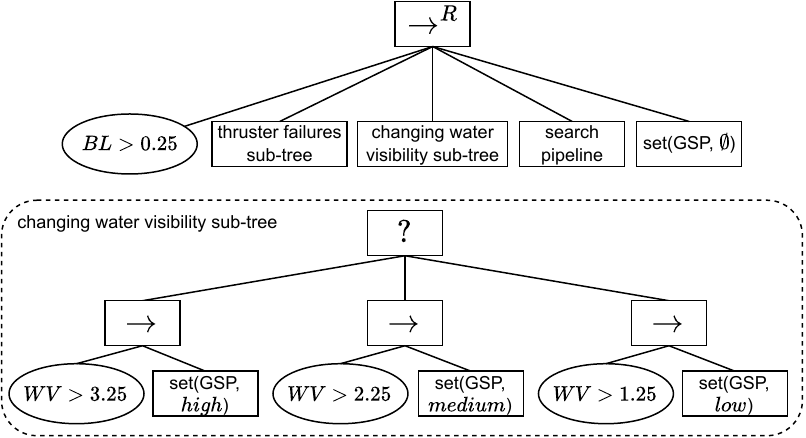}
    \hfill
    \caption{Snippet of the BT used to capture SUAVE's task and adaption logic. \emph{WV} and \emph{BL} represent the measured water visibility and battery level, respectively. \emph{GSP}, \emph{MM}, and \emph{FP} represent SUAVE's \emph{generate search path node}, \emph{maintain motion node}, and \emph{follow pipeline node}, respectively. The \emph{set} action changes the system's modes. }
    \label{fig:behaviortree_patterns}
\end{figure}

\emph{BT managing system:} 
The BT used to model SUAVE's task logic without adaptation contains 10 nodes, and the BT used for the extended use case contains 16 nodes. These values were deducted from the development effort metric for the BT managing system since they are independent of the adaptation problem. 

\Cref{fig:behaviortree_patterns} depicts the pattern used to model SUAVE's \emph{search pipeline} action and its related adaptations\footnote{The full BT can be found in the SUAVE repository.}. As can be seen, there is no separation between the task and adaptation logic, which is the main limitation of using BTs in comparison to using \acronym to model the adaptation logic. The coupling of both logics hinders the reusability of the approach as another system with the same task logic but different adaptation logic, or vice-versa, cannot reuse the existing BTs. In addition, when any changes are made to the task or adaptation problems, it will most likely require changes to parts of the BT that are not necessarily related to the changes introduced. Furthermore, it makes the modeling process more difficult as the roboticist needs to consider both problems simultaneously when modeling the BTs. 

\subsection{Development effort scalability}
\subsubsection{Experimental setup}
To evaluate how \acronym's development effort scales for more complex use cases, we analyze how the development effort of a base scenario grows with the addition of new actions and adaptations. 
The growth for adding actions and adaptations depends on the specific application.
Thus, we make the following assumptions to generalize and simplify the analysis of adding adaptation.

\begin{assumption}\label{as:1}
Every action requires one function that has only one configuration available consisting of a single component with no parameters.
\end{assumption}

\begin{assumption}\label{as:2}
Every \entity{Component} is a \entity{ROSNode} containing one \attribute{package} and one \attribute{executable} attribute; every \entity{ROSNode} has one \relationship{component configuration} with one \entity{Component Parameter}; every \relationship{function design} and \relationship{component configuration} must be related to a \relationship{constraint} and contain a \attribute{priority} attribute; and a single \entity{Quality Attribute} is defined for the whole system.
\end{assumption}

\setcounter{figure}{10}
\setcounter{subfigure}{0}
\begin{subfigure}
\setcounter{figure}{10}
\setcounter{subfigure}{0}
    \centering
    \begin{minipage}[b]{0.3\textwidth}
        \centering
        \includegraphics[width=\textwidth]{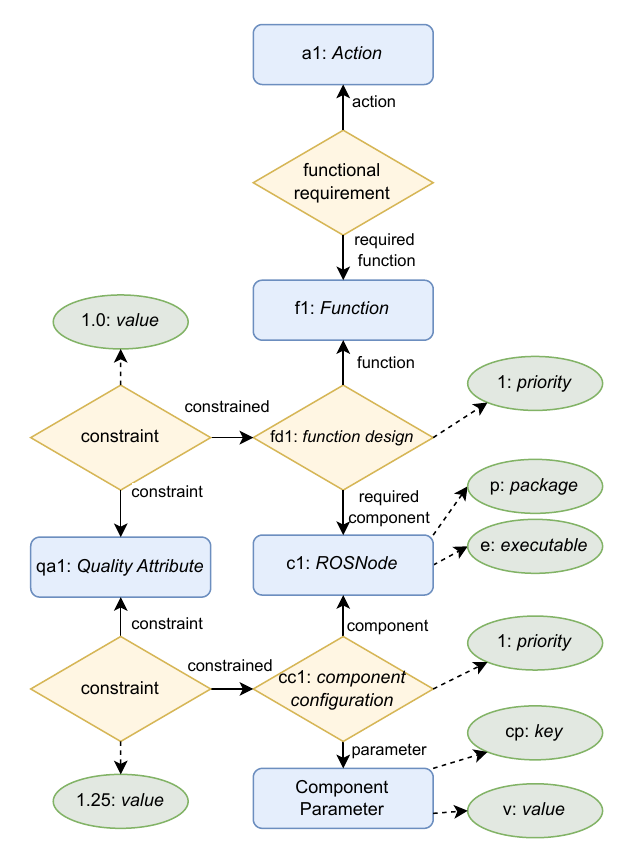}
        \caption{Base model}
        \label{fig:base_model}
    \end{minipage}  
   \hfill
\setcounter{figure}{10}
\setcounter{subfigure}{1}
    \begin{minipage}[b]{0.3\textwidth}
        \centering
        \includegraphics[width=\textwidth]{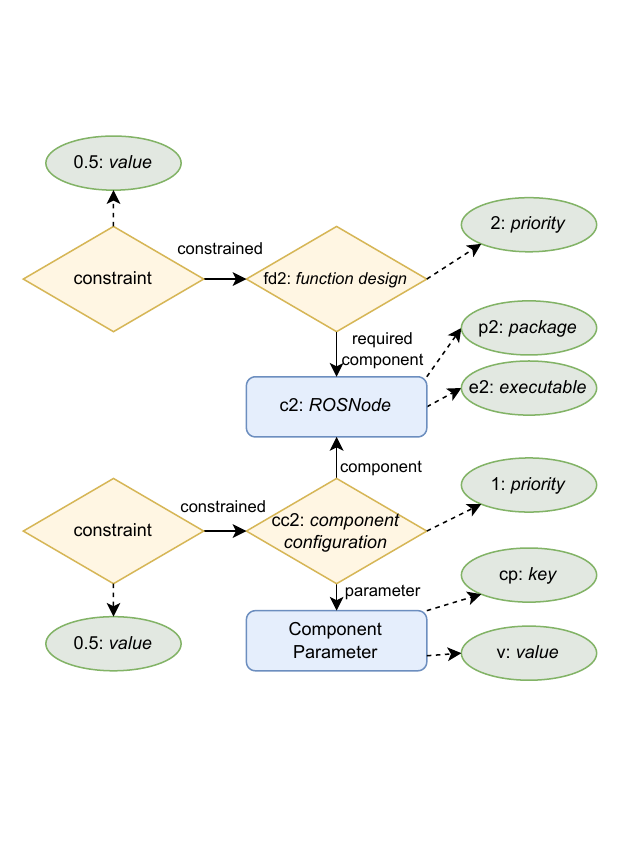}
        \caption{Structural adaptation}
        \label{fig:overhead_structural}
    \end{minipage}
    \hfill
\setcounter{figure}{10}
\setcounter{subfigure}{2}
    \begin{minipage}[b]{0.3\textwidth}
        \centering
        \includegraphics[width=\textwidth]{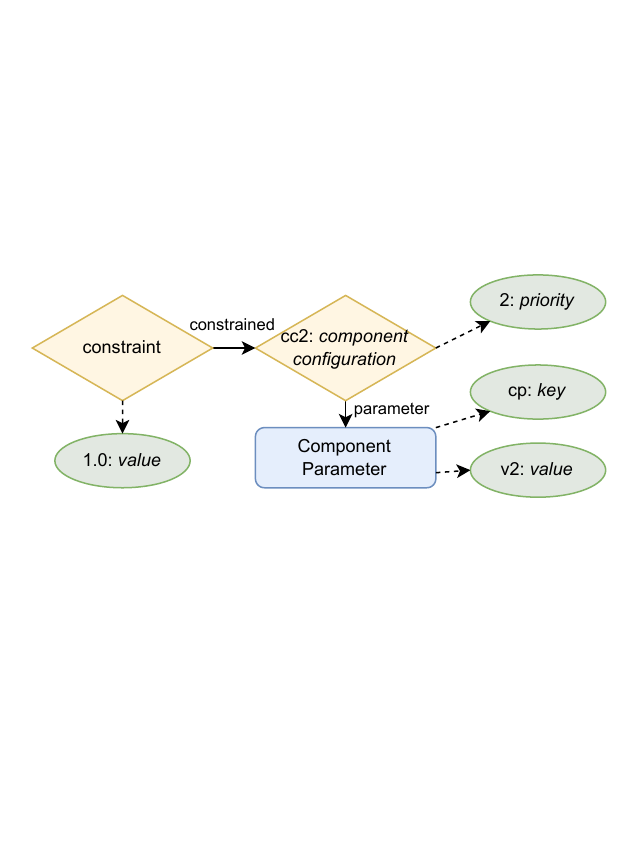}
        \caption{Parameter adaptation}
        \label{fig:overhead_parameter}
    \end{minipage}

\setcounter{figure}{10}
\setcounter{subfigure}{-1}
    \caption{Minimum knowledge required to model adaptation given Assumptions \ref{as:1} and \ref{as:2}. The name attribute is depicted as \textit{name: Entity} or \textit{name: Relationship} inside each diamond and rectangle shape, respectively.}
    \label{fig:overhead}
\end{subfigure}

\subsubsection{Results}
Consider a base scenario where the robot has only one action and complies with Assumptions~\ref{as:1} and~\ref{as:2}. The \acronym model to solve it contains $10$ elements and is depicted in \Cref{fig:base_model}.

To add structural adaptation to the base scenario (given Assumption~\mbox{\ref{as:2}}), it is necessary to add the elements depicted in \Cref{fig:overhead_structural} to the model.
This results in a minimum overhead of $6$ elements for each structural adaptation. To add parameter adaptation to the base scenario (given Assumption~\mbox{\ref{as:2}}), it is necessary to add the elements depicted in \Cref{fig:overhead_parameter} to the model.
This results in a minimum overhead of $3$ elements for each parameter adaptation.

In conclusion, given Assumptions~\ref{as:1} and~\ref{as:2}, the total overhead per action can be defined as $10 + 6*n_{sa} + 3*n_{pa}$, where $n_{sa}$ and $n_{pa}$ represent the number of structural and parameter adaptations for the action, respectively. 
This indicates that the \acronym model grows linearly with the number of actions and adaptations, which is made possible by the clear separation of the task and adaptation logic.

\section{Conclusions and future work}\label{sec:conclusion}
This work proposed \acronym, a knowledge-based solution for task-and-architecture co-adaptation in robotic systems that promotes reusability, extensibility, and composability. Reusability was achieved by proposing a knowledge model that can capture the knowledge required for TACA and using it at runtime to reason about adaptation. Extensibility and composability were achieved with an architectural design that allows ROSA's components to be stateless and self-contained. The feasibility of using \acronym in robotic systems at runtime was demonstrated by applying it in simulation to the SUAVE exemplar. \acronym's reusability was demonstrated by using it to model different self-adaptive robotic systems and showing that it can capture all relevant knowledge for adaptation necessary for these use cases.  Furthermore, \acronym's development effort and its scalability were demonstrated for the use cases presented in this paper and for a hypothetical scenario.

\acronym modular architecture has been designed to provide reuse and extensibility of the framework by future works applying self-adaptation principles in robotics architectures. For example, the current \acronym implementation supports integration with robotics architectures using planning or behavior tree solutions for the task deliberation layer, but it would be interesting to extend it to support other decision-making methods, such as state machines or Markov decision processes.

As a future work, we intend to integrate learning in ROSA. For example, machine learning methods could be explored to update at runtime the \entity{Quality Attribute} and \entity{Environmental Attribute} estimations and constraints \attribute{values}. Another possibility is learning that constraints and estimations exist without prior knowledge, i.e., learning that the relationship itself should be modeled. \acronym's interfaces to manipulate the KB at runtime could be exploited for this end.

\section*{Conflict of Interest Statement}
The authors declare that the research was conducted in the absence of any commercial or financial relationships that could be construed as a potential conflict of interest.

\section*{Author Contributions}

GRS: Conceptualization, Data Curation, Investigation, Methodology, Software, Writing – original draft, Writing – review \& editing. JP: Conceptualization, Writing – review \& editing. SLTT: Conceptualization, Writing – review \& editing. EBJ: Conceptualization, Writing – review \& editing.
CHC: Conceptualization, Writing – review \& editing.

\section*{Funding}
This work was supported by the European Union’s Horizon 2020 Framework Programme through the MSCA network REMARO (Grant Agreement No 956200).



\bibliographystyle{Frontiers-Harvard} 
\bibliography{references}

\end{document}